\documentclass[a4paper,conference]{IEEEtran}
\usepackage{soul}
\usepackage{url}
\usepackage[utf8]{inputenc}
\usepackage{graphicx}
\usepackage{amsmath}
\usepackage{amsthm}
\usepackage{booktabs}
\usepackage{algorithm}
\usepackage{algorithmic}
\usepackage{amssymb}
\usepackage{rotating}
\usepackage{framed}
\usepackage{multirow}
\usepackage{bm}
\usepackage{bbm}
\usepackage{arydshln}
\usepackage{xspace}
\usepackage{enumitem}
\usepackage[export]{adjustbox}
\usepackage{lipsum}
\usepackage{textcomp}
\usepackage{gensymb}
\usepackage{makecell}
\usepackage{array}
\usepackage{amsmath}
\usepackage{dsfont}
\usepackage{hyperref}

\def \ie {\emph{i.e.}}
\def \eg {\emph{e.g.}}
\def \etal {\emph{et al.}}

\def \etc {\emph{etc.}}

\newcommand{\tit}[1]{\smallbreak\noindent\textbf{#1.}}

\newcommand{\tinytit}[1]{\noindent\textbf{#1.}}
\newcommand{\tinytextit}[1]{\noindent\textit{#1:}}

\ifCLASSINFOpdf
\else
\fi

\begin{document}
%
\title{Spot the Difference: A Novel Task for\\Embodied Agents in Changing Environments}

\author{\IEEEauthorblockN{Federico Landi, Roberto Bigazzi, Marcella Cornia, Silvia Cascianelli, Lorenzo Baraldi, Rita Cucchiara}
\IEEEauthorblockA{University of Modena and Reggio Emilia\\
Email: \{name.surname\}@unimore.it}
}

\maketitle

\begin{abstract}
Embodied AI is a recent research area that aims at creating intelligent agents that can move and operate inside an environment. Existing approaches in this field demand the agents to act in completely new and unexplored scenes. However, this setting is far from realistic use cases that instead require executing multiple tasks in the same environment. Even if the environment changes over time, the agent could still count on its global knowledge about the scene while trying to adapt its internal representation to the current state of the environment. To make a step towards this setting, we propose \textit{Spot the Difference}: a novel task for Embodied AI where the agent has access to an outdated map of the environment and needs to recover the correct layout in a fixed time budget. To this end, we collect a new dataset of occupancy maps starting from existing datasets of 3D spaces and generating a number of possible layouts for a single environment. This dataset can be employed in the popular Habitat simulator and is fully compliant with existing methods that employ reconstructed occupancy maps during navigation.
Furthermore, we propose an exploration policy that can take advantage of previous knowledge of the environment and identify changes in the scene faster and more effectively than existing agents.
Experimental results show that the proposed architecture outperforms existing state-of-the-art models for exploration on this new setting.
\end{abstract}


%
\IEEEpeerreviewmaketitle

\section{Introduction}
\label{sec:introduction}
Imagine you have just bought a personal robot, and you ask it to bring you a cup of tea. It will start roaming around the house while looking for the cup. It probably will not come back until some minutes, as it is new to the environment. After the robot knows your house, instead, you expect it to perform navigation tasks much faster, exploiting its previous knowledge of the environment while adapting to possible changes of objects, people, and furniture positioning.
Embodied AI has recently gained a lot of attention from the research community, with amazing results in challenging tasks such as visual exploration~\cite{ramakrishnan2020exploration,bigazzi2020explore,bigazzi2022impact} and navigation~\cite{ramakrishnan2020occupancy,chaplot2020object,krantz2020beyond,landi2021multimodal}.
However, in the current setting, the environment is completely unknown to the agent as a new episode begins. We believe that this choice is not supported by real-world experience, where information about the environment can be stored and reused for future tasks.
As agents are likely to stay in the same place for long periods, such information may be outdated and inconsistent with the actual layout of the environment. Therefore, the agent also needs to discover those differences during navigation.
In this paper, we introduce a new task for Embodied AI, which we name \emph{Spot the Difference}. In the proposed setting, the agent must identify all the differences between an outdated map of the environment and its current state -- a challenge that combines visual exploration using monocular images and embodied navigation with spatial reasoning. To succeed in this task, the agent needs to develop efficient exploration policies to focus on likely changed areas while exploiting priors about objects of the environment. We believe that this task could be useful to train agents that will need to deal with changing environments.

\begin{figure}[t]
\centering
\scriptsize
\setlength{\tabcolsep}{.15em}
\begin{tabular}{cccccc}
\textbf{Original Map} & & & \multicolumn{3}{c}{\textbf{Sample Manipulated Maps}} \\
\addlinespace[0.12cm]
\includegraphics[width=0.233\linewidth]{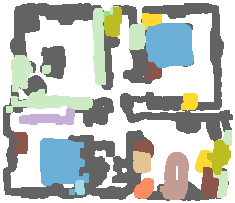} & & &
\includegraphics[width=0.233\linewidth]{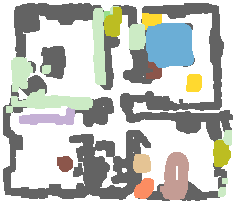} &
\includegraphics[width=0.233\linewidth]{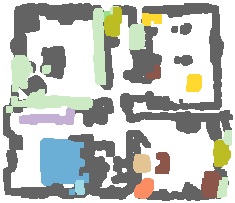} &
\includegraphics[width=0.233\linewidth]{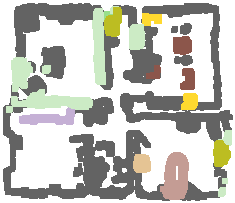} \\
\includegraphics[width=0.233\linewidth]{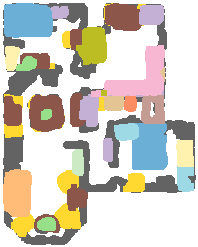} & & &
\includegraphics[width=0.233\linewidth]{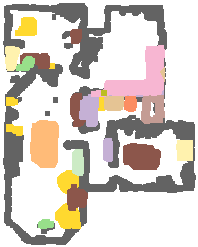} &
\includegraphics[width=0.233\linewidth]{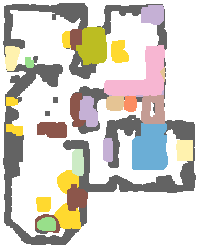} &
\includegraphics[width=0.233\linewidth]{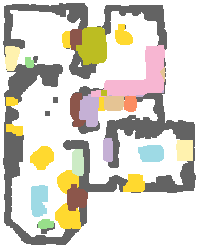} \\
\includegraphics[width=0.233\linewidth]{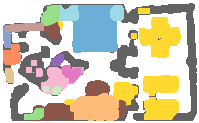} & & &
\includegraphics[width=0.233\linewidth]{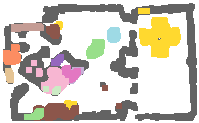} &
\includegraphics[width=0.233\linewidth]{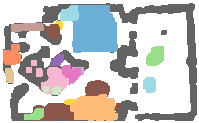} &
\includegraphics[width=0.233\linewidth]{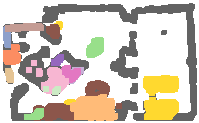} \\
\addlinespace[0.12cm]
\multicolumn{6}{c}{\includegraphics[width=0.98\linewidth]{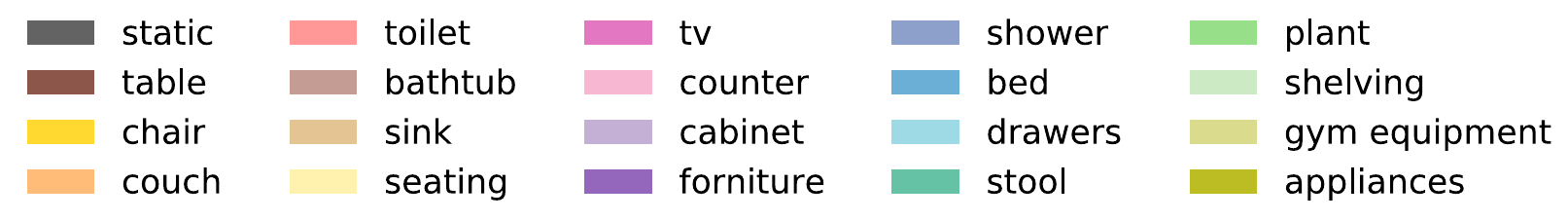}}
\end{tabular}
\caption{Generation of alternative states of an environment: original and sample manipulated semantic maps.}
\label{fig:maps}
\vspace{-0.2cm}
\end{figure}

Recent work on Embodied AI has tackled the training of embodied agents capable of navigating and locating objects~\cite{cartillier2020semantic,chaplot2020object,krantz2020beyond,wani2020multion}. One of the key factors for success in the field consists in building map representations in which knowledge about the environment can be stored while the agent proceeds~\cite{chaplot2019learning,ramakrishnan2020occupancy}. However, the dominant training and evaluation protocol involves an agent initialized from scratch that sees the environment for the first time~\cite{anderson2018evaluation}. Another line of work~\cite{chen2019learning,luperto2020robot,ramakrishnan2020exploration,karkus2021differentiable,mayo2021visual,bigazzi2022impact}, instead, introduces a mapping phase of the environment to increase the performance on both exploration and down-stream tasks. Unfortunately, if the environment changes over time, the agent needs to rebuild a full representation from scratch and cannot count on an efficient policy to update its internal representation of the environment. In this work, we simulate the natural evolution of an environment and design a specific policy to navigate in changing environments seamlessly.

Due to the high cost of 3D acquisitions from the real world, the existing datasets of photorealistic 3D spaces~\cite{chang2017matterport3d,xia2018gibson} do not contain different layouts for the same environment. In this paper, we create a reproducible set-up to generate alternative layouts for an environment. We semi-automatically build a dataset of 2D semantics occupancy maps in which the objects can be removed and rearranged while the area and the position of architectural elements do not change (Fig.~\ref{fig:maps}). In the proposed setting, the agent is deployed in an interactive 3D environment and provided with a map from our produced dataset, which represents the information retained while performing tasks in a past state of the environment.

To train agents that can deal with changing environments efficiently, we develop a novel reward function and an approach for navigation aiming at finding relevant differences between the previous layout of the environment and the current one. Our method is based on the recent Active Neural SLAM paradigm proposed in~\cite{chaplot2019learning} and~\cite{ramakrishnan2020occupancy}. Differently from previous proposals, though, it can read and update the given map to identify relevant differences in the environment in the form of their projections on the map.
Our dataset and architecture can be employed with the Habitat simulator~\cite{savva2019habitat}, a popular research platform for Embodied AI that renders photorealistic scenes and that enables seamless sim-to-real deployment of navigation agents~\cite{kadian2020sim2real,bigazzi2021out}.
Experimental results show that our approach performs better than existing state-of-the-art architectures for exploration in our newly-proposed task. We also compare with different baselines and evaluate our agent in terms of percentage of area seen, percentage of discovered differences, and metric curves at varying exploration time budgets.
The new dataset, together with our code and pretrained models, is available at this  \href{https://github.com/aimagelab/spot-the-difference}{link}.

\section{Related Work}
\label{sec:related}
Current research directions on Embodied AI for navigation agents can be categorized according to the quantity of knowledge about the environment provided to the agent prior to performing the task~\cite{anderson2018evaluation}. The first direction focuses on the scenario in which the agent is deployed in a completely new environment for which it has no prior knowledge~\cite{gupta2017cognitive,chaplot2019learning,karkus2021differentiable}. Running exploration in parallel with a target-driven navigation task resulted in an effective approach to solve the latter (\eg, object-goal navigation~\cite{chaplot2020object} and point-goal navigation~\cite{ramakrishnan2020occupancy}). 
Other directions consider the case in which the agent can exploit pre-acquired information about the environment \cite{da2018autonomously,chaplot2019embodied} when performing a navigation task. Such pre-acquired information can be either partial~\cite{savinov2018semi,sridharan2020commonsense,zhang2020diagnosing} or complete~\cite{chen2019learning,cartillier2020semantic,ramakrishnan2020exploration}. A major limitation of such approaches is that the obtained map representation is assumed to conform perfectly with the environment where the down-stream task will be performed. 

In this work, we explore a fourth direction, in which the pre-acquired map provided to the agent is incomplete or incorrect due to changes occurred in the environment over time. Common strategies to deal with changing environments entail disregarding dynamic objects as landmarks when performing SLAM~\cite{saputra2018visual,biswas2019quest} and applying local policies to avoid them when navigating~\cite{mac2016heuristic}. 
An alternative strategy is learning to predict geometric changes based on experience, as done in~\cite{nardi2020long}, where the environment is represented as a traversability graph. The main limitation of this strategy is its computational intractability when considering dense metric maps of wide areas, as in our case.

\section{Proposed Setting}
\label{sec:setting}
In the first part of this section, we introduce a new task for embodied agents, named \textit{Spot the Difference}. We then describe the newly-proposed dataset that we create to enable this setting. Finally, we propose an architecture for embodied agents to tackle the defined task.

\subsection{Spot the Difference: Task Definition}
At the beginning of an episode, the agent is spawned in a 3D environment and is given a pre-built occupancy map $M$, representing its spatial knowledge of the environment, \ie~a previous state of the environment that is now obsolete:
\begin{equation}
    M = (m_{ij}) \in [0,1], \quad 0 \leq i,j < W, 
\end{equation}
where $m_{ij}$ represents the probability of finding an obstacle at coordinates $(i,j)$.
The task entails exploring the current environment to recognize all the differences with respect to the state in which $M$ was computed, in the form of free and occupied space.
To accomplish the task, the agent should build a correct occupancy map of the current environment starting from $M$, recognizing and focusing on parts that are likely to change (\eg, the middle of wide rooms rather than tight corridors).

For every episode of \textit{Spot the Difference}, the agent is given a time budget of $T$ time-steps. At time $t=0$, the agent holds the starting map representation $M$. At each time-step $t$, the map is updated depending on the current observation to obtain $M_t$. Whenever the agent discovers a new object or a new portion of free space, the internal representation of the map changes accordingly. The goal is to gather as much information as possible about changes in the environment by the end of the episode. To measure the agent performance, we compare the final map $M_T$ produced by the agent with the ground-truth occupancy map $M^*$.
In this sense, the paradigm we adopt is the one of knowledge reuse starting from partial knowledge.

\begin{table}[t!]
\centering
\caption{Number of manipulated maps generated per dataset split.}
\label{tab:dataset_detail}
\setlength{\tabcolsep}{.4em}
\resizebox{\linewidth}{!}{
\begin{tabular}{lccccc}
\toprule
\textbf{Dataset Split} & & \textbf{Semantic Classes} & \textbf{Scans} & \textbf{Generated SOMs} & \textbf{Episodes} \\
\midrule
MP3D Train  & & 42 & 58 & 2070 & $\approx 4.5\text{M}$ \\
MP3D Validation  & & 42 & 9 & 160 & 320 \\
MP3D Test  & & 42 & 14 & 260 & 610 \\
Gibson Validation  & & 20 & 5 & 130 & 450 \\
\bottomrule
\end{tabular}
}
\vspace{-0.2cm}
\end{table}

\subsection{Dataset Creation} \label{sec:dataset}

\begin{figure*}[t!]
    \centering
    \includegraphics[width=\linewidth]{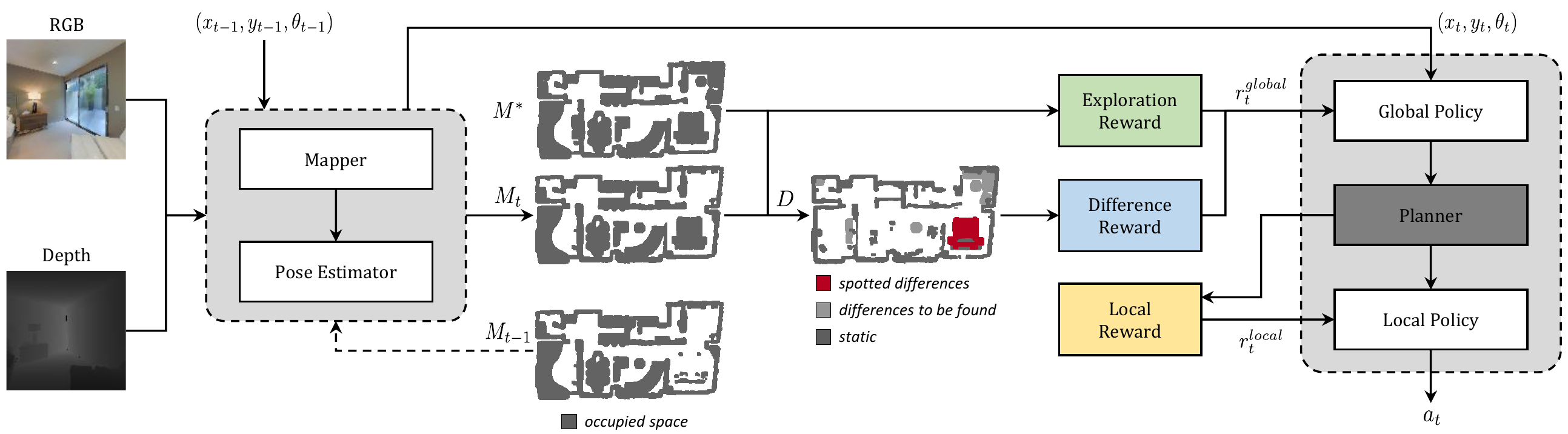}
    \caption{Overview of the proposed approach for navigation in changing environments.}
    \label{fig:method}
    \vspace{-0.2cm}
\end{figure*}

\tinytit{Semantic Occupancy Map}
Given a 3D environment, we place the agent in a free navigable location with heading $\theta=0\degree$ (facing eastward).
We assume that the input consists of a depth image and a semantic image and that the camera intrinsics $K$ are known.
To build the Semantic Occupancy Map (SOM) of an environment, we project each semantic pixel of the acquired scene into a 2-dimensional top-down map:
given a pixel with image coordinates $(i,j)$ and depth value $d_{i,j}$, we first recover its coordinates $(x,y,z)$ with respect to the agent position. 
Then, we compute the corresponding $(u,v)$ pixel in map through an orthographic projection, using the information about the agent position and heading:
\begin{equation}
    \begin{bmatrix}
        x \\ y \\ z
    \end{bmatrix}
     = d_{i,j} K^{-1}
     \begin{bmatrix}
         i \\ j \\ 1
     \end{bmatrix}
     \text{,}\quad\text{and}\quad
     \begin{bmatrix}
         u \\ v \\ 0 \\ 1
     \end{bmatrix}
     = P_v
     \begin{bmatrix}
         x \\ y \\ z \\ 1
     \end{bmatrix} .
\end{equation}
We perform the same operation after rotating the agent by $\Delta_\theta=30\degree$ until we perform a span from $0\degree$ to $180\degree$. To cover the whole scene, we repeat this procedure placing the agent at a distance of $0.5m$ from the previous capture point, following the axis directions. The agent elevation is instead kept fixed. During this step, we average the results of subsequent observations of overlapping portions of space.

After the acquisition, we obtain a SOM with $C$ channels, where each pixel corresponds to a $5cm \times 5cm$ portion of space in the 3D environment. For each channel $c \in \{0, ..., C\}$, the map values represent the probability that the corresponding portion of space is occupied by an object of semantic class $c$.

\tinytit{Multiple Semantic Occupancy Maps for the Same Environment}
The SOMs obtained in the previous step can be seen as one possible layout for the corresponding 3D environments.
In order to create a dataset with different states (\ie~different layouts) of the same environment, instead of manipulating the real-world 3D scenes (changing the furniture position, removing chairs, \etc), we propose to modify the SOM to create a set of plausible and different layouts for the environment.

First, we isolate the objects belonging to each semantic category by using an algorithm for connected component labeling~\cite{grana2010optimized,bolelli2019spaghetti,allegretti2019optimized}. Then, we sample a subset of objects to be deleted from the map and a subset of objects to be re-positioned in a different free location of the map. During sampling, we consider categories that have a high probability of being displaced or removed in the real world and ignore non-movable semantic categories such as \textit{fireplaces}, \textit{columns}, and \textit{stairs}.
After this step, we obtain a new SOM representing a possible alternative state for the environment, which could be very different from the one in which the 3D acquisition was taken. Sample manipulated maps can be found in Fig.~\ref{fig:maps}.

\tinytit{Dataset Details}
To generate alternative SOMs, we start from the Matterport 3D (MP3D) dataset of spaces~\cite{chang2017matterport3d}, which comprises $90$ different building scans, and is enriched with dense semantic annotations.
We consider each floor in the building and compute the SOM for that floor. For each map, we create 10 alternative versions of that same environment. In this step, we discard the floors that have few semantic objects (\eg,~empty rooftops) or that are not fully navigable by the agent. As a result, we retain $249$ floors belonging to $81$ different buildings, thus generating a total of $2490$ different semantic occupancy maps for these floors.  Finally, we split the dataset into train, validation, and test subsets.

As an additional test bed, we also build a set of out-of-domain maps ($13$ floors from $5$ spaces) taken from the Gibson dataset~\cite{xia2018gibson}, enriched with semantic annotations from~\cite{armeni20193d}, and manipulated as done for the MP3D dataset. For each SOM, multiple episodes are generated by selecting different starting points.
More information about our dataset can be found in Table~\ref{tab:dataset_detail} and in the supplementary material.

\subsection{Agent Architecture}
Our model for embodied navigation in changing environments comprises three major components: a mapper module, a pose estimator, and a navigation policy (which, in turn, consists of a global policy, a planner, and a local policy). An overview of the proposed architecture is shown in Fig.~\ref{fig:method} and described below, while additional details can be found in the supplementary material.
Although the data we provide is enriched with semantic labels, our agent does not make use of such information directly. This is in line with current state-of-the-art architectures for embodied exploration that we choose as  competitors.

\tinytit{Mapper}
The mapper module takes as inputs an RGB observation $o^r_t$ and the corresponding depth image $o^d_t$,  representing the first-person view of the agent at time-step $t$, and outputs the agent-centric occupancy map $v_t$ of a $V\times V$ region in front of the camera. Each pixel in $v_t$ corresponds to a $25mm \times25mm$ portion of space and consists of two channels containing the probability of that cell being occupied and explored, respectively.
As a first step, we encode $o^r_t$ using the first two blocks of ResNet-18 pre-trained on ImageNet, followed by a three-layer CNN. We project the depth image $o^d_t$ using the camera intrinsics~\cite{chen2019learning} and obtain a preliminary map for the visible occupancy. We name the obtained feature representations $\hat{o}^r_t$ and $\hat{o}^d_t$, respectively. We then encode the two feature maps using a U-Net~\cite{ronneberger2015u}:
\begin{equation}
    f_\mu(\hat{o}^r_t, \hat{o}^d_t) = \text{U-Net}_\text{enc}(\hat{o}^r_t, \hat{o}^d_t, \mu) , 
\end{equation}
and decode the $2 \times V \times V$ matrix of probabilities as:
\begin{equation}
    v_t = \sigma(\text{U-Net}_\text{dec}(f_\mu(\hat{o}^r_t, \hat{o}^d_t), \phi)) ,
\end{equation}
where $\mu$ and $\phi$ represent the learnable parameters in the U-Net encoder and decoder, respectively, and $\sigma$ is the sigmoid activation function.
The computed agent-centric occupancy map $v_t$ is then registered in the global occupancy map $M_{t-1}$ coming from the previous time-step to obtain $M_t$. 
To that end, we use a geometric transformation to project $v_t$ in the global coordinate system, for which we need a triple $(x,y,\theta)$ corresponding to the agent position and heading in the environment. This triple is estimated by a specific component that tracks the agent displacements across the environment, as discussed in the following paragraph.

\tinytit{Pose Estimator}
The agent can move across the environment using three actions: \textit{go forward 0.25m}, \textit{turn left 10\degree}, \textit{turn right 10\degree}. Since each action may produce a different outcome because of physical interactions with the environment (\eg,~bumping into a wall) or noise in the actuation system, the pose estimator is used to estimate the real displacement made at every time-step.
We estimate the agent displacement $(\Delta x_t,\Delta y_t,\Delta \theta_t)$ at time-step $t$ by using two consecutive RGB and depth observations, as well as the agent-centric occupancy maps $(v_{t-1}, v_t)$ computed by the mapper at $t-1$ and $t$.
The actual agent position $(x_t,y_t,\theta_t)$ is computed iteratively as:
\begin{equation}
    (x_t,y_t,\theta_t) = (x_{t-1},y_{t-1},\theta_{t-1}) + (\Delta x_t,\Delta y_t,\Delta \theta_t).
    \label{eq:pose}
\end{equation}
We assume that the agent starting position is the triple $(x_0,y_0,\theta_0) = (0,0,0)$.

\tinytit{Global Policy, Planner, and Local Policy}
The sampling of atomic actions for the exploration relies on a three-component hierarchical policy.
The first component is the global policy, which samples a long-term global goal on the map.
The global policy outputs a probability distribution over discretized locations of the global map. We sample the global goal from this distribution and then transform it in $(x,y)$ global coordinates.
The second component is a planner module, which employs the A* algorithm to decode a local goal on the map. The local goal is an intermediate point, within \textit{0.25m} from the agent, along the trajectory towards the global goal.
The last element of our navigation module is the local policy, which decodes the series of atomic actions taking the agent towards the local goal.
In particular, the local policy is an RNN decoding the atomic action $a_t$ to execute at every time-step.
The reward $r^{local}_t$ given to the local policy is proportional to the reduction in the Euclidean distance $d$ between the agent position and the current local goal:
\begin{equation}
r^{local}_t = d_t - d_{t-1}.
\end{equation}

Following the hierarchical structure, a global goal is sampled every N time-steps. A new local goal is computed if a new global goal is sampled, if the previous local goal is reached, or if the local goal location is known to be not traversable.

\tinytit{Exploiting Past Knowledge for Efficient Navigation}
The global policy is trained using a two-term reward. The first term encourages exhaustive exploration and is proportional either to the increase of area-coverage~\cite{chen2019learning} or to the increase of anticipated map accuracy as in~\cite{ramakrishnan2020occupancy}. Intuitively, the agent strives to maximize the portion of the seen area and thus maximizes the knowledge gathered during exploration. 
Moreover, since we consider a setting where a significant amount of knowledge is already available to the agent, we add a reward term to guide the agent towards meaningful points of the map. These correspond to the coordinates where major changes are likely to happen.

Given the occupancy map of the agent at time $t$, $M_t$, the true occupancy map for the same environment $M^{*}$, and a time budget of $T$ time-steps for exploration, we aim to minimize the following, for $0<t\leq T$:
\begin{equation}
    D = \sum \mathds{1} [M_t \neq M^{*} ]
\end{equation}
In other words, we want to maximize the number of pixels in the online reconstructed map $M_t$ that the agent correctly shifts from free to occupied (and vice-versa) during exploration. This leads to the reward term for difference discovery:
\begin{equation}
    r_{\text{diff}} = \sum \mathds{1} [M_t = M^{*} ] - \sum \mathds{1} [M_{t-1} = M^{*}].
\end{equation}

The proposed reward term is designed to encourage navigation towards areas in the map that are more likely to contain meaningful differences (\eg,~rooms containing more objects that can be displaced or removed from the scene). Additionally, an agent trained with this reward will tend to avoid difficult spots that are likely to produce a mismatch in terms of the predicted occupancy maps. This is because errors in the mapping phase would result in a negative reward. 
\begin{table*}[t]
\centering
\caption{Experimental results on MP3D test set. The agent incorporating the proposed reward term for discovered differences outperforms the competitors on the main metrics for the novel Spot the Difference task.
} 
\label{tab:mp3d_results}
\setlength{\tabcolsep}{.4em}
\resizebox{\linewidth}{!}{
\begin{tabular}{lc ccccccccc c ccccccccc}
\toprule
 & & \multicolumn{9}{c}{\textbf{Estimated Localization}} & & \multicolumn{9}{c}{\textbf{Oracle Localization}}\\
\cmidrule{3-11} \cmidrule{13-21} 
 & & $\mathsf{Seen [\%]}$ & $\mathsf{Acc.}$ & $\mathsf{IoU}_{+}$ & $\mathsf{IoU}_{-}$ & $\mathsf{IoU}$ & $\mathsf{mAcc.}$ & $\mathsf{mIoU}_{+}$ & $\mathsf{mIoU}_{-}$ & $\mathsf{mIoU}$  & & $\mathsf{Seen [\%]}$ & $\mathsf{Acc.}$ & $\mathsf{IoU}_{+}$ & $\mathsf{IoU}_{-}$ & $\mathsf{IoU}$ & $\mathsf{mAcc.}$ & $\mathsf{mIoU}_{+}$ & $\mathsf{mIoU}_{-}$ & $\mathsf{mIoU}$\\
\midrule
\textbf{OccAnt}   & & 52.1 & 26.2 & 13.4 & 6.1 & 11.5 & 51.1 & 19.1 & 8.3 & 15.8
                            & & 49.0 & 35.6 & 26.5 & 16.1 & 24.8 & 77.8 & 49.2 & 23.6 & 43.2 \\
\midrule
\textbf{DR}   & & 49.4 & 29.3 & 15.3 & 8.7 & 13.9 & 59.7 & 23.1 & 11.9 & 20.2
                        & & 48.6 & 37.4 & 27.2 & 18.4 & 26.5 & 80.1 & 49.8 & 27.4 & 45.8 \\
\textbf{AR} & & 43.8 & 30.6 & 19.7 & 12.9 & 18.8 & 72.5 & 36.8 & 18.4 & 32.7
            & & 43.6 & 32.5 & 23.2 & 17.5 & 23.0 & 78.7 & 47.5 & 26.7 & 44.5 \\
\textbf{CR} & & \textbf{53.2} & 33.1 & 18.1 & 9.6 & 16.1 & 65.2 & 26.4 & 12.7 & 22.6
            & & \textbf{52.8} & 39.2 & 29.6 & 18.8 & 28.0 & 78.5 & 51.0 & 26.6 & 45.7 \\
\midrule
\textbf{AR+DR}  & & 51.4 & 34.5 & 20.9 & 12.0 & 19.3 & 71.5 & 33.9 & 16.2 & 30.0 
                & & 51.4 & 37.8 & 27.3 & 18.0 & 26.2 & 79.3 & 48.9 & 25.8 & 44.4 \\
\textbf{CR+DR}  & & 52.3 & \textbf{37.8} & \textbf{24.2} & \textbf{14.8} & \textbf{22.7} & \textbf{76.2} & \textbf{39.1} & \textbf{19.8} & \textbf{34.8}
                & & 51.8 & \textbf{40.3} & \textbf{29.2} & \textbf{19.2} & \textbf{28.1} & \textbf{82.1} & \textbf{50.4} & \textbf{26.9} & \textbf{46.2} \\               
\bottomrule
\end{tabular}
}
\vspace{-0.2cm}
\end{table*}

To train our model, we combine a reward promoting exploration and the more specific reward on found differences to exploit semantic clues in the environment:
\begin{equation}
    \label{eq:rglobal}
    r^{global}_t = \beta_1 r_{\text{exp}} + \beta_2 r_{\text{diff}}
\end{equation}
where $r_\text{exp}$ is the reward term encouraging task-agnostic exploration (such as coverage-based or anticipation-based rewards, as described in the next section), and $\beta_1$ and $\beta_2$ are two coefficients weighing the importance of the two elements.

\section{Experiments and Results}
\label{sec:experiments}
In this section, we detail our experimental setting and show experimental results for our new proposed task. Further analysis can be found in the supplementary material.

\tinytit{Evaluation Metrics}
To evaluate the performance in \textit{Spot the Difference}, we consider three main classes of metrics. First, we consider the percentage of navigable area in the environment seen by the agent during the episode ($\mathsf{Seen[\%]}$). Then, we evaluate the percentage of elements that have been correctly detected as changed in the occupancy map ($\mathsf{Acc.}$) and the pixel-wise Intersection over Union for the \textit{changed} occupancy map elements ($\mathsf{IoU}$). Besides, we evaluate the task as a two-class problem and compute the $\mathsf{IoU}$ score for objects that were added in place of free space ($\mathsf{IoU_+}$) and for objects that were deleted during the map creation ($\mathsf{IoU_-}$). In addition, to evaluate the performance independently from the exploration capability, we propose to compute the metrics only on the portion of space that the agent actually visited
($\mathsf{mAcc.}$, $\mathsf{mIoU}$, $\mathsf{mIoU_+}$, and $\mathsf{mIoU_-}$)

\tinytit{Implementation Details}
We conduct our experiment using Habitat~\cite{savva2019habitat}, a popular platform for Embodied AI in photo-realistic indoor environments~\cite{xia2018gibson,chang2017matterport3d}.
The agent observations are $128 \times 128$ RGB-D images from the environment.
The learning algorithm adopted for training is PPO~\cite{schulman2017proximal}. The learning rate is $10^{-3}$ for the mapper and $2.5\times10^{-4}$ for the other modules.
Every model is trained for $\approx6.5$M frames using Adam optimizer~\cite{kingma2015adam}.
A global goal is sampled every $N=25$ time-steps. The local and global policies are updated, respectively, every $N$ and $20\times N$ time-steps, and the mapper is updated every $4\times N$ time-steps.
The size of the local map is $V=101$, while the global map size is set to $W=2001$ for the MP3D dataset and to $W=961$ for the Gibson dataset. The global policy action space size $G$ is $240$.
The reward coefficients $\{\beta_1, \beta_2\}$ are set to $\{1, 10^{-2}\}$ and $\{1, 10^{-1}\}$ when the exploration reward is based on coverage and anticipation reward, respectively.
The length of each episode is fixed to $T=1000$ time-steps.

\tinytit{Competitors and Baselines}
We consider the following competitors and variants of the proposed method on two different setups: one where the agent position is predicted by the agent (as in Eq.~\ref{eq:pose}), and one where it has access to oracle coordinates:

\tinytextit{Difference Reward (DR)} an exploration policy that maximizes the correctly predicted changes between $M$ and $M^*$. This corresponds to setting $\beta_1=0$ and $\beta_2=1$ in Eq.~\ref{eq:rglobal}.

\tinytextit{Coverage Reward (CR)} an agent that explores the environment with an exploration policy that maximizes the covered area and builds the occupancy map as it goes, as in~\cite{ramakrishnan2020occupancy}. 

\tinytextit{Anticipation Reward (AR)} an agent that explores the environment with an exploration policy that maximizes the covered area and the correctly anticipated values in the occupancy map built as it goes,  from~\cite{ramakrishnan2020occupancy}.
Our proposed approach consists of an agent trained with the combination of the difference reward with the coverage reward (\textit{CR+DR}) or with the anticipation reward (\textit{AR+DR}).

\tinytextit{Occupancy Anticipation (OccAnt)} we also compare with the agent presented by Ramakrishnan \etal~\cite{ramakrishnan2020occupancy} using the available pre-trained models, referenced to as \textit{OccAnt}. Note that \textit{OccAnt} was trained on the Gibson dataset for the standard exploration task and without any prior map. Thus, it is not directly comparable with the other methods considered. We include it to gain insights into the performance of an off-the-shelf state-of-the-art agent on our task.

\begin{figure}[t]
\centering
\scriptsize
\setlength{\tabcolsep}{.2em}
\begin{tabular}{cccc}
& & \textbf{Cumulative $\mathsf{Acc.}$} & \textbf{Cumulative $\mathsf{IoU}$} \\
\rotatebox{90}{\parbox[t]{1.2in}{\hspace*{\fill}\textbf{Estimated Localization}\hspace*{\fill}}} & & 
\includegraphics[width=0.463\linewidth]{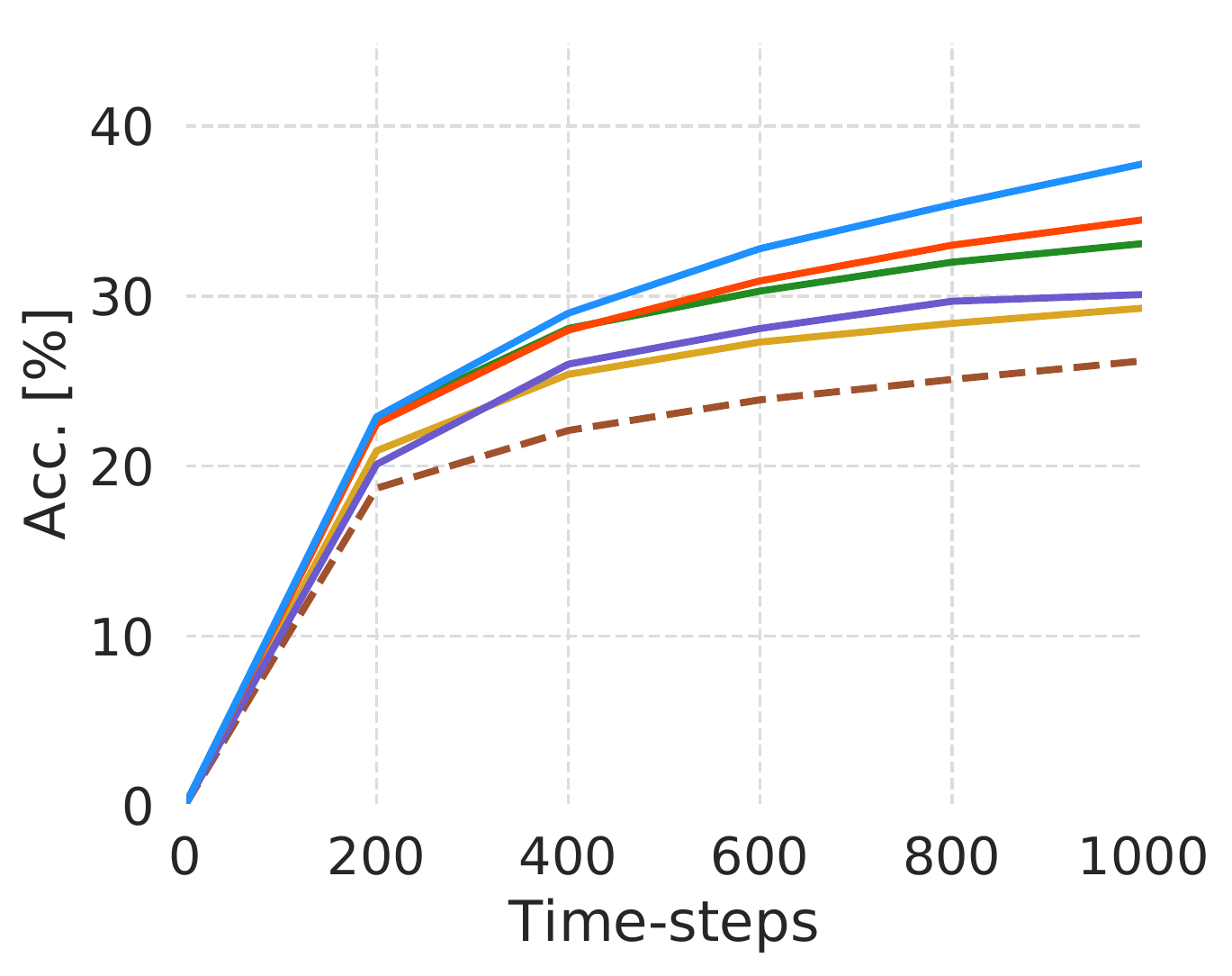} & \includegraphics[width=0.463\linewidth]{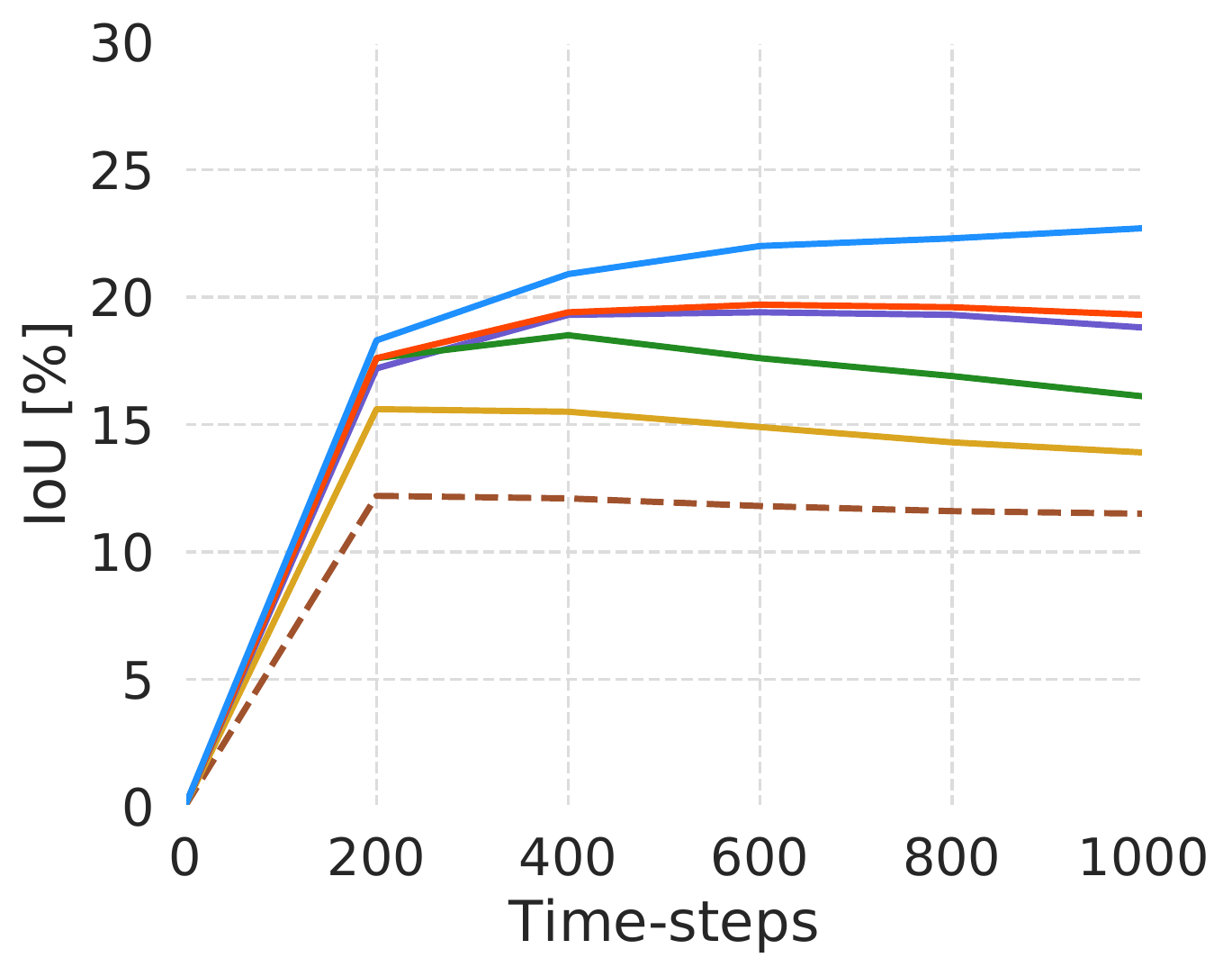} \\ 
\rotatebox{90}{\parbox[t]{1.2in}{\hspace*{\fill}\textbf{Oracle Localization}\hspace*{\fill}}} & &
\includegraphics[width=0.463\linewidth]{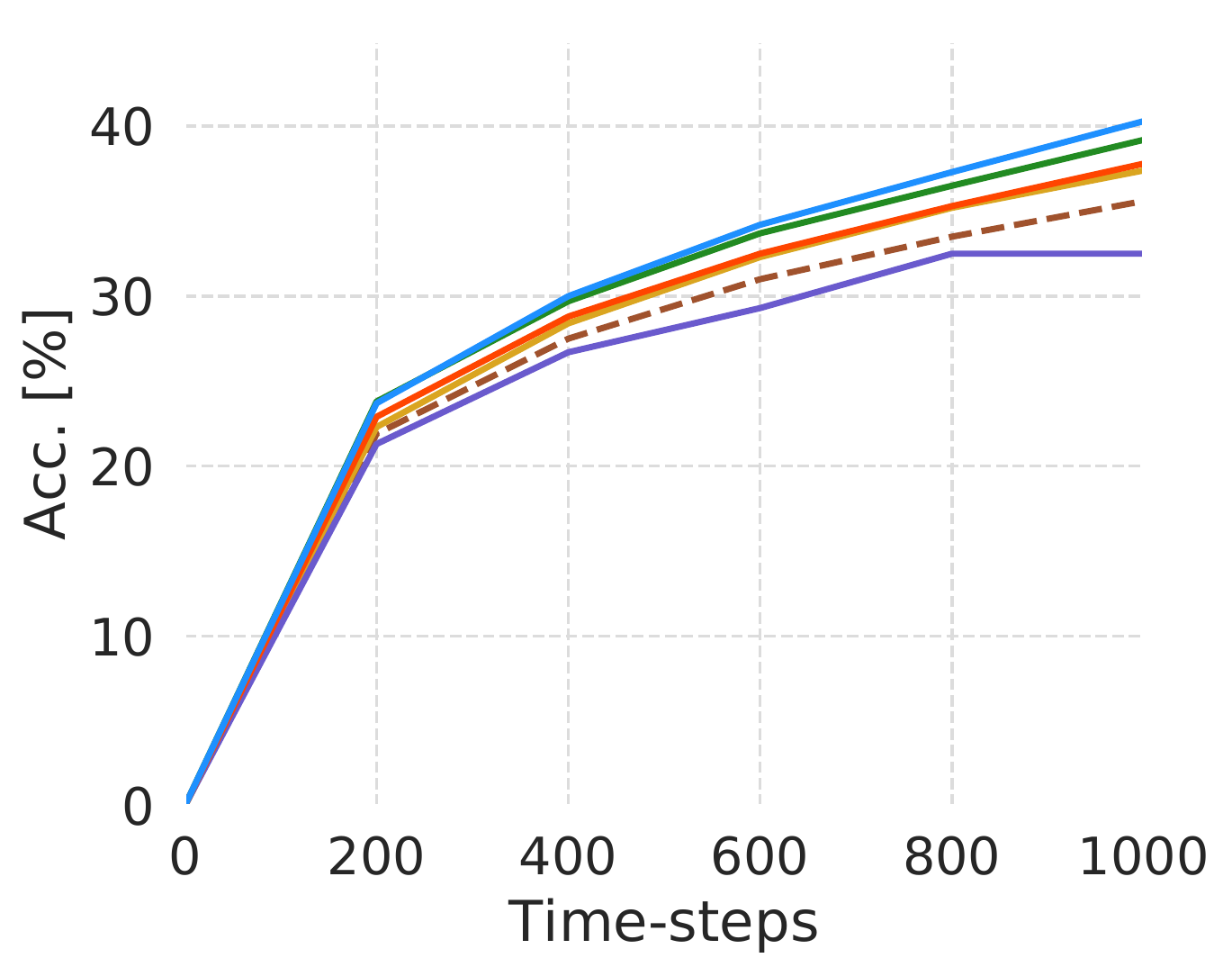} & \includegraphics[width=0.463\linewidth]{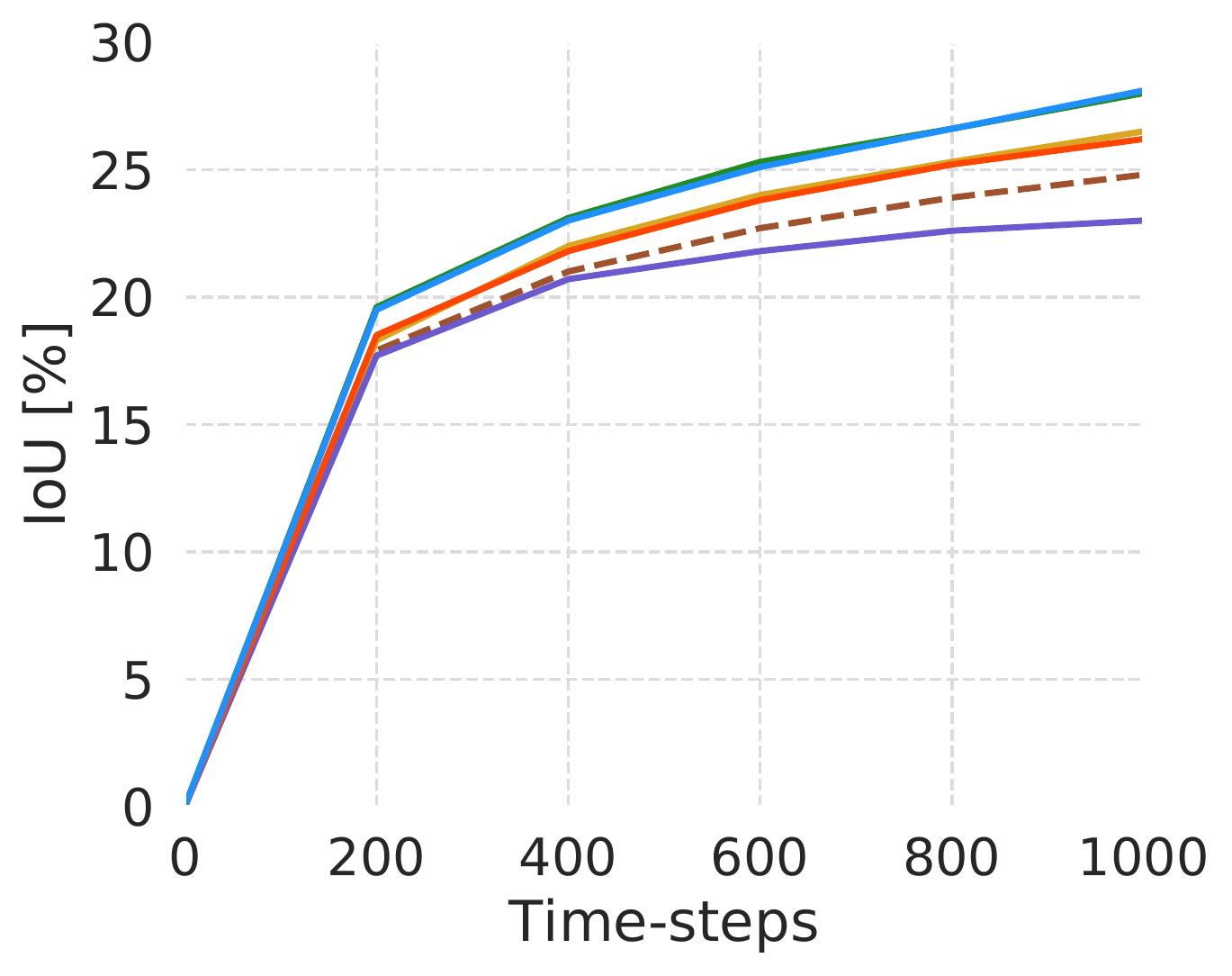} \\
& & \multicolumn{2}{c}{\includegraphics[width=0.93\linewidth]{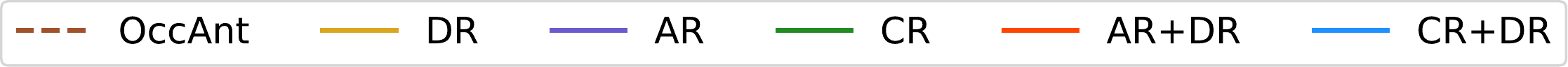}} \\
\end{tabular}
\caption{Value of accuracy and IoU for the different models at varying time-steps on the MP3D test set.}
\label{fig:curves}
\vspace{-0.2cm}
\end{figure}

\tinytit{Results on MP3D dataset}
As a first testbed, we evaluate the different agents on the MP3D \textit{Spot the Difference} test set.
We report the results for this experiment in Table~\ref{tab:mp3d_results}.

We observe that the agent combining a reward based on coverage and our reward based on the differences in the environment (\textit{CR+DR}) performs best on all the pixel-based metrics and places second in terms of percentage of seen area. It is worth noting that, even if the results in terms of the area seen are not as high as the ones obtained by the \textit{CR} agent, the addition of our Difference Reward helps the agent to focus on more relevant parts, and thus, it can discover more substantial differences. Additionally, predictions are more accurate and more precise, as indicated by the $4.7\%$ and $6.6\%$ improvements in terms of $\mathsf{Acc.}$ and $\mathsf{IoU}$ with respect to the \textit{CR} competitor. Instead, a reward based on differences alone is not sufficient to promote good exploration. In fact, although the \textit{DR} agent outperforms the \textit{CR} and \textit{AR} agents on some metrics, our reward alone does not provide as much improvement as when combined with rewards encouraging exploration (as for \textit{CR+DR} and \textit{AR+DR}).

\begin{table}[t]
\centering
\caption{Experimental results on Gibson validation set.
} 
\label{tab:gibson_results}
\setlength{\tabcolsep}{.4em}
\resizebox{\linewidth}{!}{
\begin{tabular}{lc ccccccccc}
\toprule
& & \multicolumn{9}{c}{\textbf{Estimated Localization}} \\
\cmidrule{3-11}
& & $\mathsf{Seen [\%]}$ & $\mathsf{Acc.}$ & $\mathsf{IoU}_{+}$ & $\mathsf{IoU}_{-}$ & $\mathsf{IoU}$ & $\mathsf{mAcc.}$ & $\mathsf{mIoU}_{+}$ & $\mathsf{mIoU}_{-}$ & $\mathsf{mIoU}$ \\
\midrule
\textbf{OccAnt} & & 86.2 & 49.8 & 11.9 & 7.2 & 10.4 & 58.0 & 12.3 & 7.5 & 10.8 \\
\midrule
\textbf{DR}   & & \textbf{86.2} & 53.2 & 13.2 & 8.5 & 11.7 & 63.7 & 13.9 & 8.8 & 12.3 \\
\textbf{AR} & & 75.3 & 51.5 & 21.4 & 16.6 & 20.4 & 72.7 & 25.8 & 17.3 & 23.3 \\
\textbf{CR} & & 85.9 & 57.6 & 16.7 & 11.9 & 15.4 & 71.3 & 18.6 & 12.3 & 16.7 \\
\midrule
\textbf{AR+DR} & & 83.4 & 58.7 & 20.0 & 14.9 & 19.0 & 75.8 & 23.0 & 15.6 & 21.1 \\
\textbf{CR+DR} & & 82.1 & \textbf{60.1} & \textbf{24.0} & \textbf{19.0} & \textbf{23.1} & \textbf{78.5} & \textbf{27.8} & \textbf{19.9} & \textbf{25.9} \\
\bottomrule
\end{tabular}
}
\vspace{-0.2cm}
\end{table}

Even in the oracle localization setup,
the \textit{CR+DR} agent achieves the best results. Interestingly, the gap with the \textit{CR} agent decreases to $1.1\%$ and $0.1\%$ in terms of $\mathsf{Acc.}$ and $\mathsf{IoU}$, respectively. This is because our \textit{CR+DR} agent learns to sample trajectories that can be performed more efficiently and without accumulating a high positioning error. For this reason, the performance boost given by the oracle localization is lower. For both setups, our \textit{CR+DR} agent outperforms the state-of-the-art \textit{OccAnt} agent for exploration on all the metrics.

Finally, in Fig.~\ref{fig:curves}, we plot different values of $\mathsf{Acc.}$ and $\mathsf{IoU}$ over different time-steps during the episodes. This way, we can evaluate the whole exploration trend, and not only its final point.
We can observe that the proposed models incorporating the difference reward outperform the competitors. In particular, the \textit{CR+DR} agent scores first by a significant margin. The performance gap can be noticed even in the first half of the episode and tends to grow with the number of steps.

\begin{figure}[t]
\centering
\scriptsize
\setlength{\tabcolsep}{.2em}
\resizebox{\linewidth}{!}{
\begin{tabular}{cccc}
\textbf{Starting Map} &\textbf{CR} & \textbf{CR+DR} & \textbf{Ground-truth Map} \\
\addlinespace[0.12cm]
\includegraphics[width=0.21\linewidth]{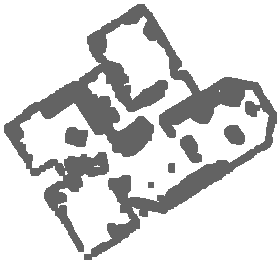} &
\includegraphics[width=0.21\linewidth]{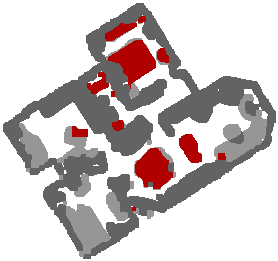} &
\includegraphics[width=0.21\linewidth]{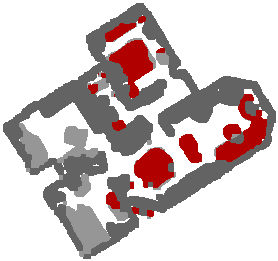} & 
\includegraphics[width=0.21\linewidth]{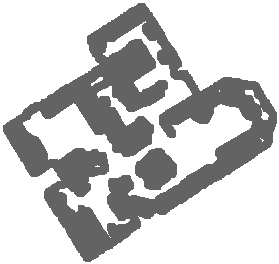} \\
\includegraphics[width=0.21\linewidth]{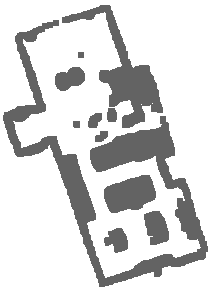} &
\includegraphics[width=0.21\linewidth]{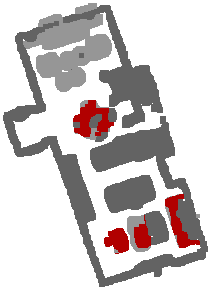} &
\includegraphics[width=0.21\linewidth]{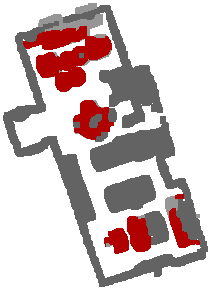} & 
\includegraphics[width=0.21\linewidth]{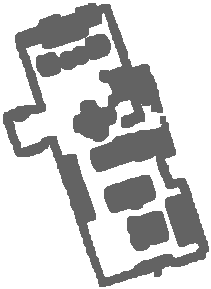} \\
\includegraphics[width=0.21\linewidth]{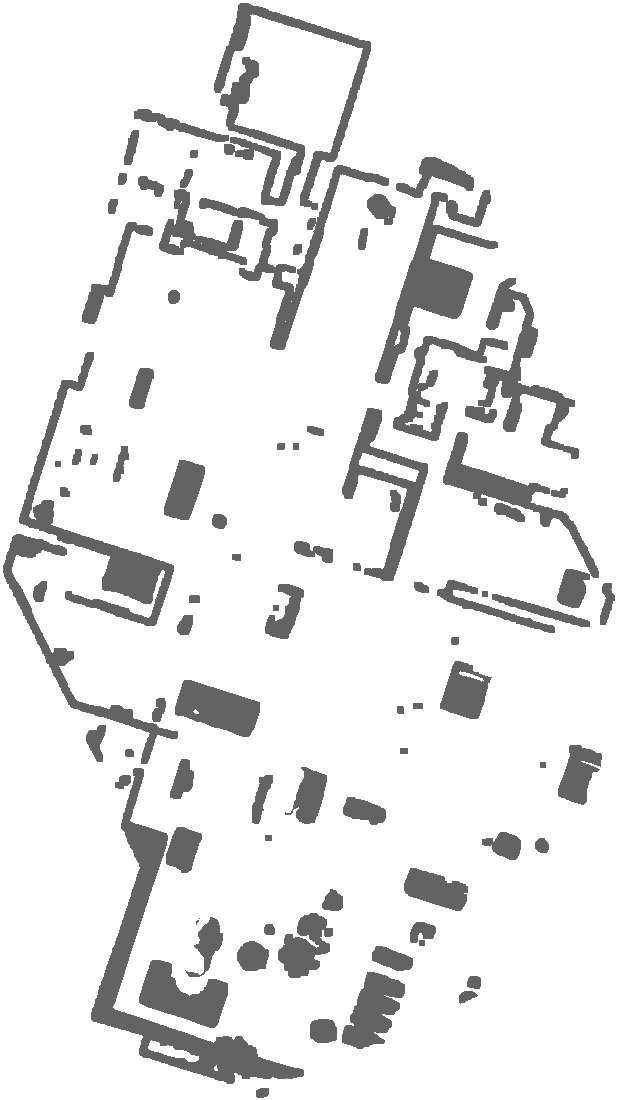} &
\includegraphics[width=0.21\linewidth]{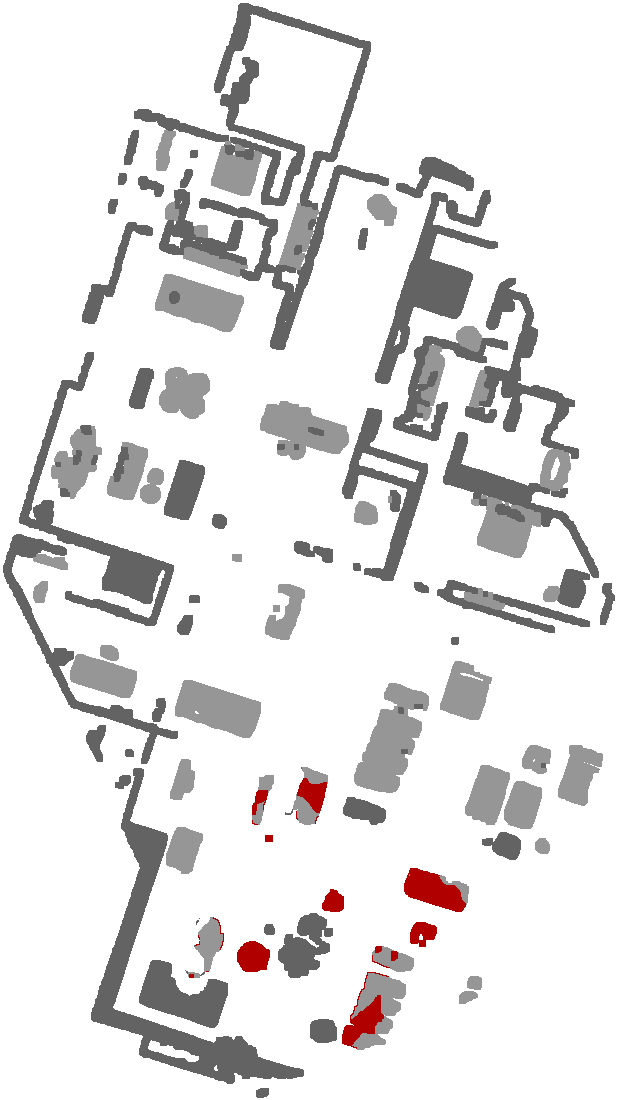} &
\includegraphics[width=0.21\linewidth]{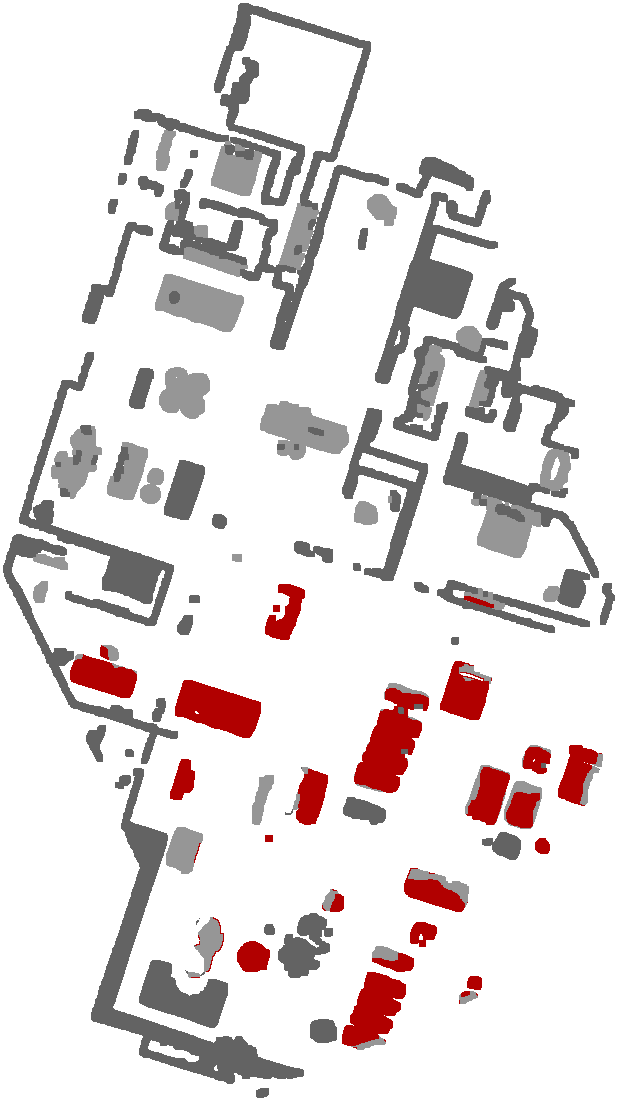} & 
\includegraphics[width=0.21\linewidth]{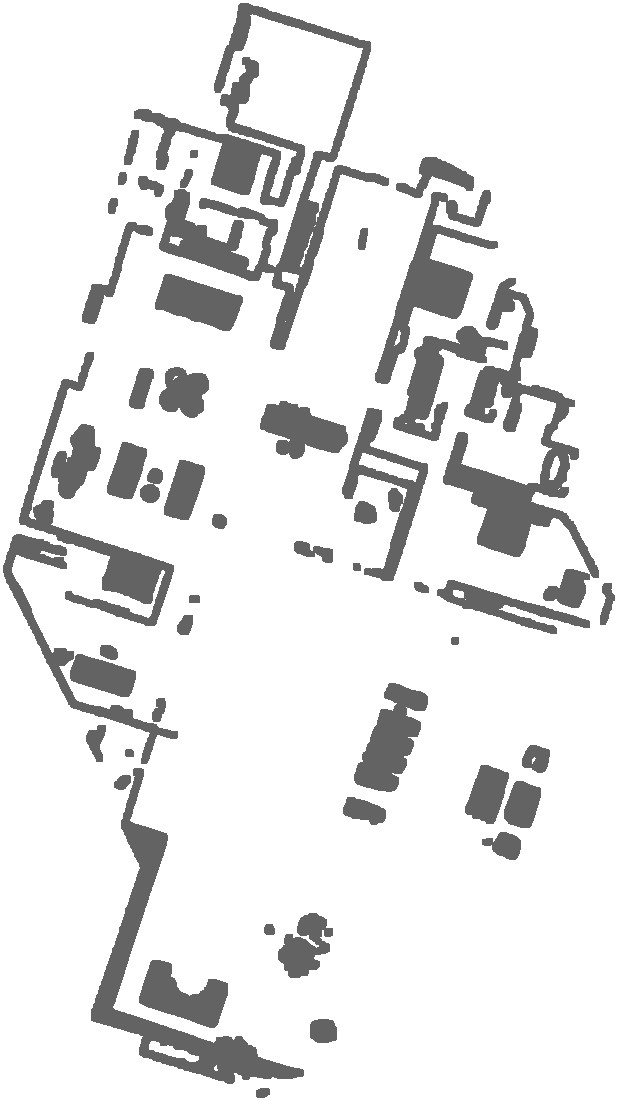} \\
\addlinespace[0.12cm]
\multicolumn{4}{c}{\includegraphics[width=0.95\linewidth]{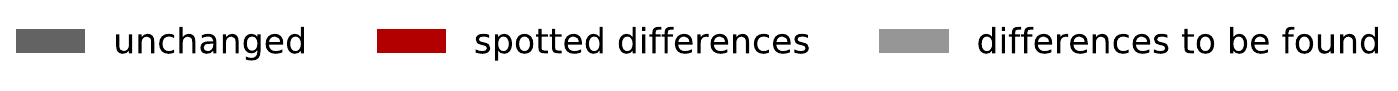}}
\end{tabular}
}
\caption{Qualitative results comparing the performances of the CR and CR+DR agents for different episodes.}
\label{fig:differences}
\vspace{-0.2cm}
\end{figure}

\tinytit{Results on Gibson dataset}
The environments from the Gibson dataset~\cite{xia2018gibson} are generally smaller than those in MP3D, and thus, they can be explored more easily and exhaustively. We report the results for this experiment in Table~\ref{tab:gibson_results}.
Also in this experiment, the \textit{CR+DR} agent performs best on all the metrics but the percentage of the area seen. Although \textit{CR+DR} explores $3.8\%$ of the environment less than the \textit{CR} agent, it still overcomes the competitor by $2.5\%$ and $7.7\%$ in terms of $\mathsf{Acc.}$ and $\mathsf{IoU}$. The \textit{AR+DR} agent is the second-best in terms of $\mathsf{Acc.}$. The \textit{OccAnt} agent, instead, is competitive in terms of area seen but achieves low $\mathsf{Acc.}$ and $\mathsf{IoU}$ metrics.

\tinytit{Qualitative Results}
In Fig.~\ref{fig:differences}, we report some qualitative results. Starting from the left-most column, we present the starting map given to the agent as the episode begins, the results achieved by the \textit{CR} agent, those of the proposed \textit{CR+DR} agent, and the ground-truth map. The differences that the agents have correctly identified during the episode are highlighted in red. As it can be seen, the \textit{CR+DR} agent can identify more differences than the \textit{CR} counterpart, even in small environments (top row). As the size of the environments grows (bottom row), the performance gap increases and the \textit{CR+DR} agent outperforms its competitor.

\section{Conclusion}
\label{sec:conclusion}
In this work, we proposed \emph{Spot the Difference}: a new task for navigation agents in changing environments. In this novel setting, the agent has to find all variations that occurred in the environment with respect to an outdated occupancy map. Since current datasets of 3D spaces do not account for such variety, we collected a new dataset containing different layouts for the same environment. 
We tested two state-of-the-art exploration agents on this task and proposed a novel reward term to encourage the discovery of meaningful information during exploration. The proposed agent outperforms the competitors and can identify changes in the environment more efficiently. We believe that the results presented in this paper motivate further research on this new proposed setting for Embodied AI.


\section*{Acknowledgment}
This work has been supported by the ``European Training Network on PErsonalized Robotics as SErvice Oriented applications'' (PERSEO) MSCA-ITN-2020 project (G.A. 955778).

\bibliographystyle{IEEEtran}
\bibliography{bibliography}

\appendix

\subsection{Additional Implementation Details}


\tinytit{Semantic Classes Division}
The generation of semantic maps for each floor of each scene produces $2001 \times 2001 \times 43$ maps for the MP3D dataset and $961 \times 961 \times 21$ maps for the Gibson dataset. The last channel of every map registers the explorable space, so it is ignored for the creation of the dataset and is concatenated, as it is, to the manipulated map obtained at the end of the semi-automatic dataset creation process. 

We divide the semantic channels of the maps depending on the possible actions performable on the connected components in that channel. We identify four types of classes: \textit{No Operation}, \textit{Removal}, \textit{Displacement}, and \textit{Overlap Removal}. A list of semantic categories with their classification is reported in Table~\ref{tab:semantic_classes_mp3d} for the MP3D dataset and in Table~\ref{tab:semantic_classes_gibson} for the Gibson dataset.
\textit{No Operation} classes are left untouched, and correspond to non movable objects, such as \textit{wall}, \textit{stairs} and \textit{columns}; the connected component of the \textit{Removal} classes can be removed; those in the \textit{Displacement} classes can be either removed or relocated in other free spaces in the environment; and \textit{Overlap Removal} components are removed if connected components removed or displaced in other channels overlap with them, \eg,~if a \textit{sofa} is removed, every instance of \textit{cushion} overlapping with that \textit{sofa} will be removed as well because it is supposed to be on top of it.

In Fig.~\ref{fig:sup_maps}, we report some additional examples of manipulated semantic maps with relative difference maps obtained by applying our semi-automatic procedure.

\tit{Train/Val/Test Splits Division}
We use the same scene partitioning adopted by the existing datasets for embodied exploration and PointGoal navigation on Matterport3D and Gibson Tiny~\cite{chang2017matterport3d,xia2018gibson}. 

\tit{Episode Creation}
For the creation of the episodes of our dataset we use the starting positions of the exploration dataset for MP3D, and of the PointGoal navigation dataset for Gibson Tiny. After the episodes located in floors with few semantic objects or that are not fully navigable by the agent are discarded, we associate one of the alternative versions of the ground-truth semantic map to each episode. For the validation and test splits of the MP3D dataset and the validation split of the Gibson dataset we create new episodes with random sampled starting positions so that the number of episodes on every floor is at least $10$ and fix the number of episodes per floor to a multiple of $10$. We report a detailed list of scans, selected floors and number of episodes per scan in Tables~\ref{tab:sup_mp3dtrain},~\ref{tab:sup_mp3dval},~\ref{tab:sup_mp3dtest}, and~\ref{tab:sup_gibsonval}. 


\tinytit{Pose Estimator}
The pose estimator takes as input two consecutive RGB and depth observations, consisting in two pairs $(o^r_{t-1}, o^d_{t-1})$ and $(o^r_t, o^d_t)$. Additionally, it accepts as input the agent-centric occupancy maps $(v_{t-1}, v_t)$ computed by the mapper at $t-1$ and $t$. 
For each modality, we encode information using a CNN followed by a fully-connected layer. We call these intermediate representations $\bar{o}^r_t$, $\bar{o}^d_t$, and $\bar{v}_t$. 
Then, we compute a first estimate of the relative displacement in terms of $(x,y,\theta)$ coordinates and heading for each modality:
\begin{equation}
    g(\star) = W_1 \text{max}(W_2\star + b_2,0) + b_1 ,
    \label{eq:gstar}
\end{equation}
with $\star \in \{\bar{o}^r_t, \bar{o}^d_t, \bar{v}_t \}$.
We stack the vectors computed in Eq.~\ref{eq:gstar} to obtain a $3 \times 3$ matrix $G$.
Finally, we compute the agent displacement at time-step $t$ $(\Delta x_t,\Delta y_t,\Delta \theta_t)$ as:
\begin{equation}
    (\Delta x_t,\Delta y_t,\Delta \theta_t) = \sum_{i=1}^{3}{\alpha_i \cdot G_i},
\end{equation}
\begin{equation}
    \alpha_i = \text{softmax}(\text{MLP}_i([\bar{o}^r_t, \bar{o}^d_t, \bar{v}_t])),
\end{equation}
where $G_i$ indicates the $i$-th row of the $G$ matrix, 
MLP is a three-layer fully-connected network, and $[\cdot, \cdot, \cdot]$ denotes tensor concatenation.

\tit{Global Policy}
An enriched occupancy map $M_t^+ \in [0,1]^{4\times W \times W}$ is obtained by stacking the occupancy map, the map of visited states, and the one-hot representation of the agent location $(x_t, y_t)$. Then, we compute two versions of $M_t^+$: one by cropping the map to an agent-centered $G \times G$ area, and the other by max-pooling the map to the same spatial resolution. The $8$-channel tensor obtained by concatenating these two versions of $M_t^+$ is fed to the global policy.
The global policy consists of a CNN that outputs a probability distribution over the $G\times G$ global action space. We sample the global goal from this distribution, and then transform it in $(x,y)$ global coordinates.

\begin{table*}[t!]
\centering
\caption{Experimental results on MP3D validation set.} 
\label{tab:mp3d_val_results}
\setlength{\tabcolsep}{.35em}
\resizebox{\linewidth}{!}{
\begin{tabular}{lc ccccccccc c ccccccccc}
\toprule
 & & \multicolumn{9}{c}{\textbf{Estimated Localization}} & & \multicolumn{9}{c}{\textbf{Oracle Localization}}\\
\cmidrule{3-11} \cmidrule{13-21} 
 & & $\mathsf{Seen [\%]}$ & $\mathsf{Acc.}$ & $\mathsf{IoU}_{+}$ & $\mathsf{IoU}_{-}$ & $\mathsf{IoU}$ & $\mathsf{mAcc.}$ & $\mathsf{mIoU}_{+}$ & $\mathsf{mIoU}_{-}$ & $\mathsf{mIoU}$  & & $\mathsf{Seen [\%]}$ & $\mathsf{Acc.}$ & $\mathsf{IoU}_{+}$ & $\mathsf{IoU}_{-}$ & $\mathsf{IoU}$ & $\mathsf{mAcc.}$ & $\mathsf{mIoU}_{+}$ & $\mathsf{mIoU}_{-}$ & $\mathsf{mIoU}$\\
\midrule
\textbf{OccAnt}$^\dagger$   & & 50.3 & 24.1 & 9.6 & 5.8 & 8.5 & 49.5 & 13.9 & 7.0 & 11.6 
                            & & 47.5 & 33.5 & 21.5 & 17.4 & 20.6 & 76.0 & 41.6 & 23.0 & 20.1 \\
\midrule
\textbf{DR} & & 48.0 & 28.7 & 11.9 & 7.3 & 10.7 & 59.7 & 17.3 & 9.0 & 14.7
            & & 44.7 & 35.2 & 22.3 & 19.0 & 22.1 & 79.4 & 42.4 & 25.0 & 38.2 \\
\textbf{AR} & & 40.1 & 29.4 & 15.9 & 13.0 & 15.7 & 71.3 & 28.8 & \textbf{16.8} & 25.9
            & & 39.2 & 31.0 & 19.2 & 17.7 & 19.4 & 77.6 & 38.9 & 24.8 & 36.0 \\
\textbf{CR} & & \textbf{53.2} & 34.7 & 14.6 & 8.7 & 12.9 & 64.1 & 20.3 & 10.3 & 17.0
            & & \textbf{52.2} & \textbf{42.6} & \textbf{26.9} & 19.8 & \textbf{25.4} & 80.4 & \textbf{46.4} & 25.2 & \textbf{40.2} \\
\midrule
\textbf{AR+DR}  & & 48.7 & 34.9 & 16.6 & 11.9 & 15.7 & 71.7 & 26.7 & 14.2 & 22.9 
                & & 47.6 & 38.0 & 22.6 & 19.3 & 22.4 & \textbf{81.3} & 42.0 & 25.4 & 37.5 \\
\textbf{CR+DR}  & & 50.1 & \textbf{38.6} & \textbf{20.1} & \textbf{14.1} & \textbf{19.1} & \textbf{75.4} & \textbf{31.8} & \textbf{16.8} & \textbf{27.6}
                & & 50.0 & 40.9 & 25.1 & \textbf{19.9} & 24.3 & 80.8 & 43.8 & \textbf{26.2} & 39.2 \\
\bottomrule
\end{tabular}
}
\vspace{-0.2cm}
\end{table*}

\begin{table}[t!]
\centering
\caption{Experimental results on Gibson validation set using the oracle localization setup..} 
\label{tab:gibson_oracle_results}
\setlength{\tabcolsep}{.35em}
\resizebox{\linewidth}{!}{
\begin{tabular}{lc ccccccccc}
\toprule
 & & \multicolumn{9}{c}{\textbf{Oracle Localization}}\\
\cmidrule{3-11}
 & & $\mathsf{Seen [\%]}$ & $\mathsf{Acc.}$ & $\mathsf{IoU}_{+}$ & $\mathsf{IoU}_{-}$ & $\mathsf{IoU}$ & $\mathsf{mAcc.}$ & $\mathsf{mIoU}_{+}$ & $\mathsf{mIoU}_{-}$ & $\mathsf{mIoU}$  \\
\midrule
\textbf{OccAnt} & & 81.6 & 60.1 & \textbf{32.1} & 21.2 & 29.2 & 78.7 & \textbf{39.6} & 22.2 &  \textbf{34.1} \\
\midrule
\textbf{DR} & & \textbf{86.1} & \textbf{65.2} & 30.1 & \textbf{24.1} & \textbf{29.9} & 81.1 & 36.0 & 25.2 & 33.8 \\
\textbf{AR} & & 74.1 & 53.8 & 27.9 & 21.9 & 27.2 & 77.0 & 35.4 & 23.5 & 32.7 \\
\textbf{CR} & & 84.0 & 62.2 & 30.6 & 22.1 & 28.8 & 79.5 & 36.1 & 23.3 & 32.8 \\
\midrule
\textbf{AR+DR}  & & 83.2 & 63.2 & 29.6 & 23.8 & 29.1 & \textbf{81.6} & 35.8 & 25.1 & 33.7 \\
\textbf{CR+DR}  & & 82.6 & 63.8 & 30.3 & \textbf{24.1} & 29.5 & \textbf{81.6} & 36.1 & \textbf{25.5} & 34.0 \\
\bottomrule
\end{tabular}
}
\end{table}


\tinytit{List of Hyperparameters}
In Table~\ref{tab:hyperparams}, we list the hyperparameters used to train our agents. These hyperparameters are shared among all the agents. Agent-depending values are specified in the main paper.

\tit{Training setup}
We train every agent using 16 NVIDIA V100 GPUs in parallel, distributed on 4 different nodes with 4 GPUs each. On every node, we run 8 Habitat environments, for a total of 32 environments in parallel. To coordinate the interaction among nodes and train our models, we use PyTorch. Each train took about 48 hours using this setup.

\subsection{Additional Experimental Results}

\tinytit{Results on MP3D Validation Set}
We report the quantitative results on the MP3D validation set in Table~\ref{tab:mp3d_val_results}.
Discussion for these experiments is analogous to the one presented for the experiments on the MP3D test set in the main paper.

\tit{Results on Gibson Validation Set with Oracle Localization}
We report the quantitative results on the Gibson validation set using the oracle localization setup in Table~\ref{tab:gibson_oracle_results}. In this setting, the agent using only the proposed difference reward (\textit{DR}) performs the best on almost all the metrics. We can conclude that, for small environments, and given an optimal localization system, our reward alone is sufficient to surpass the competitors on \textit{Spot the Difference}.

\begin{figure}[t]
\centering
\scriptsize
\setlength{\tabcolsep}{.2em}
\begin{tabular}{cccc}
& & \textbf{Cumulative $\mathsf{Acc.}$} & \textbf{Cumulative $\mathsf{IoU}$} \\
\rotatebox{90}{\parbox[t]{1.2in}{\hspace*{\fill}\textbf{Estimated Localization}\hspace*{\fill}}} & & 
\includegraphics[width=0.463\linewidth]{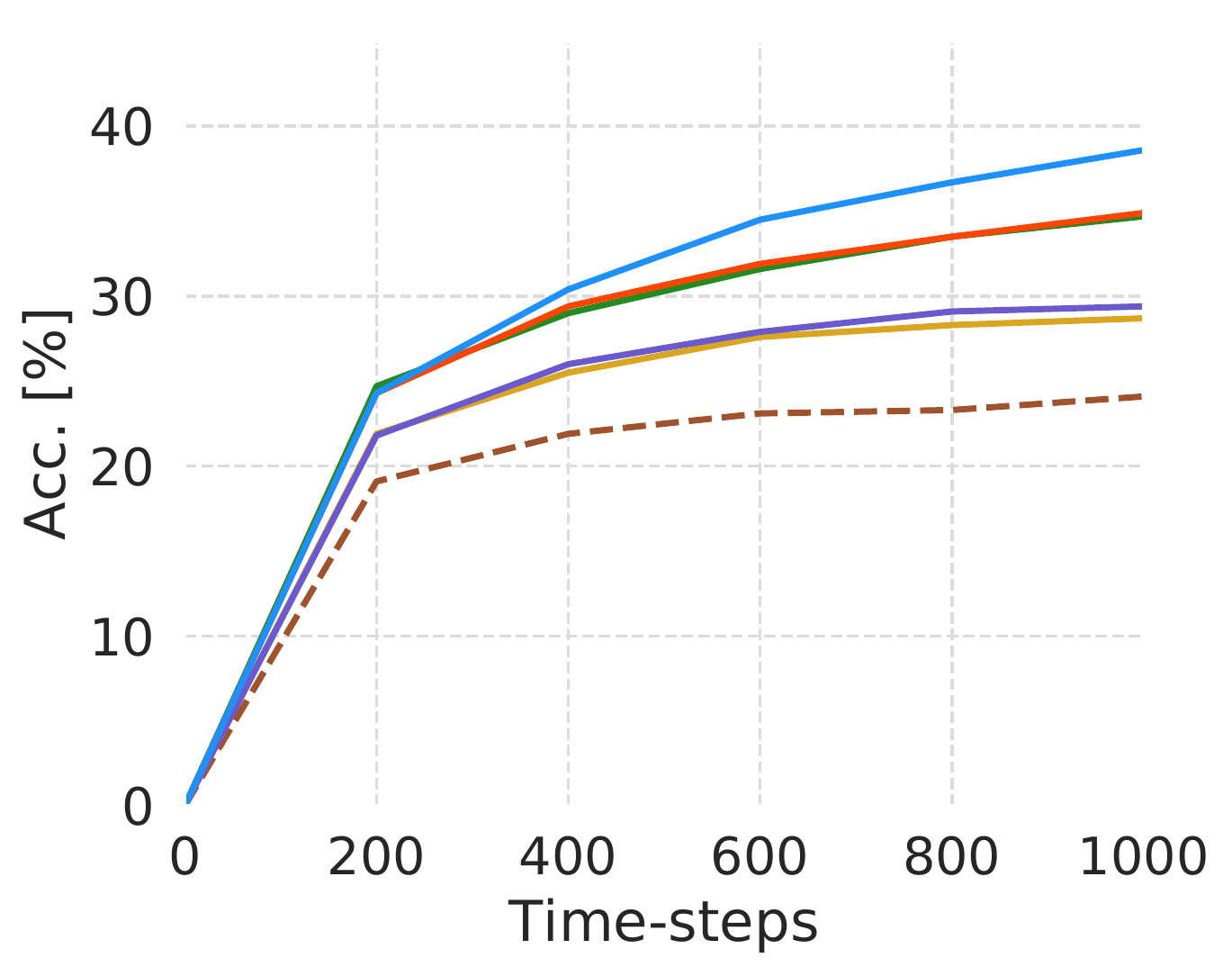} & \includegraphics[width=0.463\linewidth]{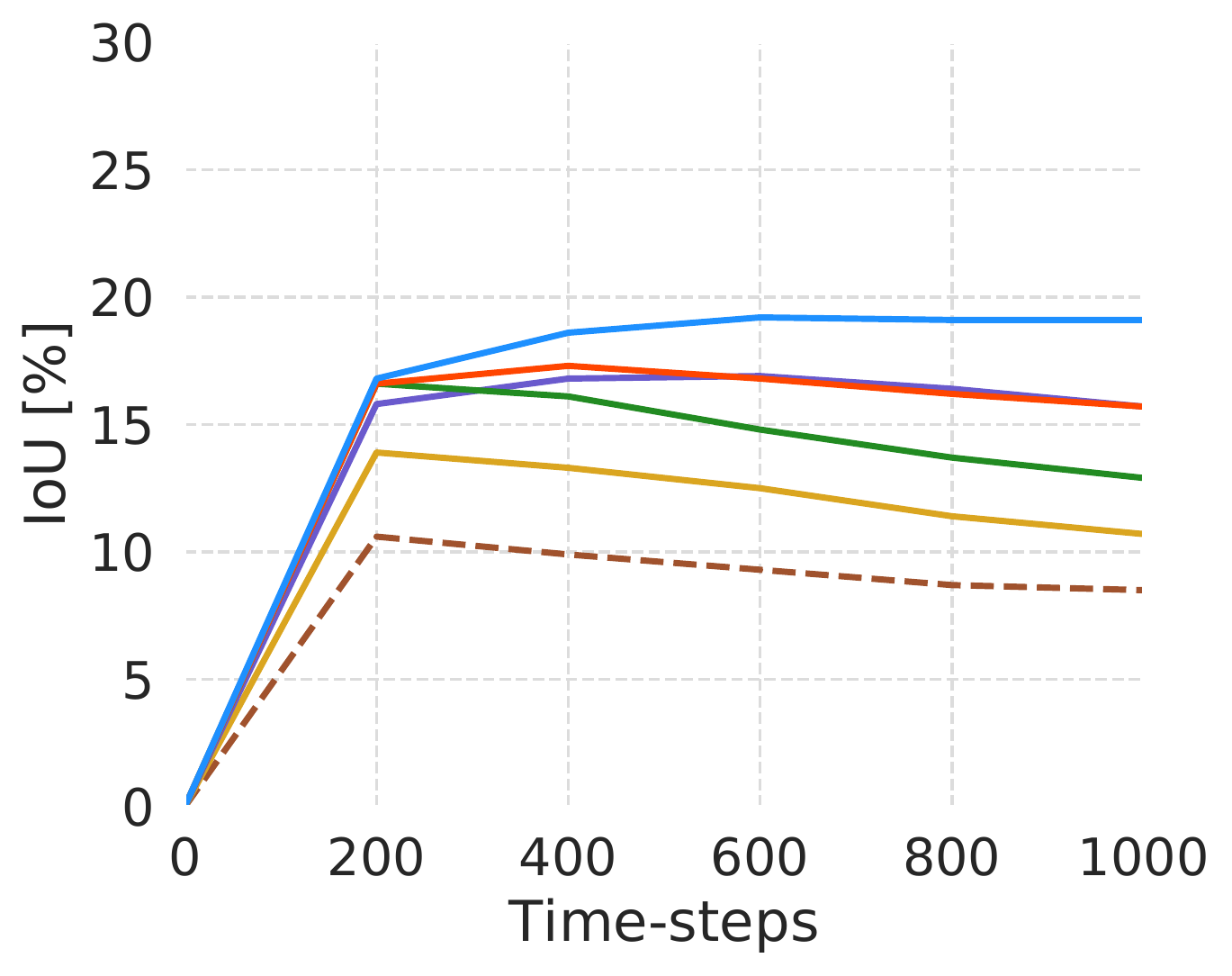} \\ 
\rotatebox{90}{\parbox[t]{1.2in}{\hspace*{\fill}\textbf{Oracle Localization}\hspace*{\fill}}} & &
\includegraphics[width=0.463\linewidth]{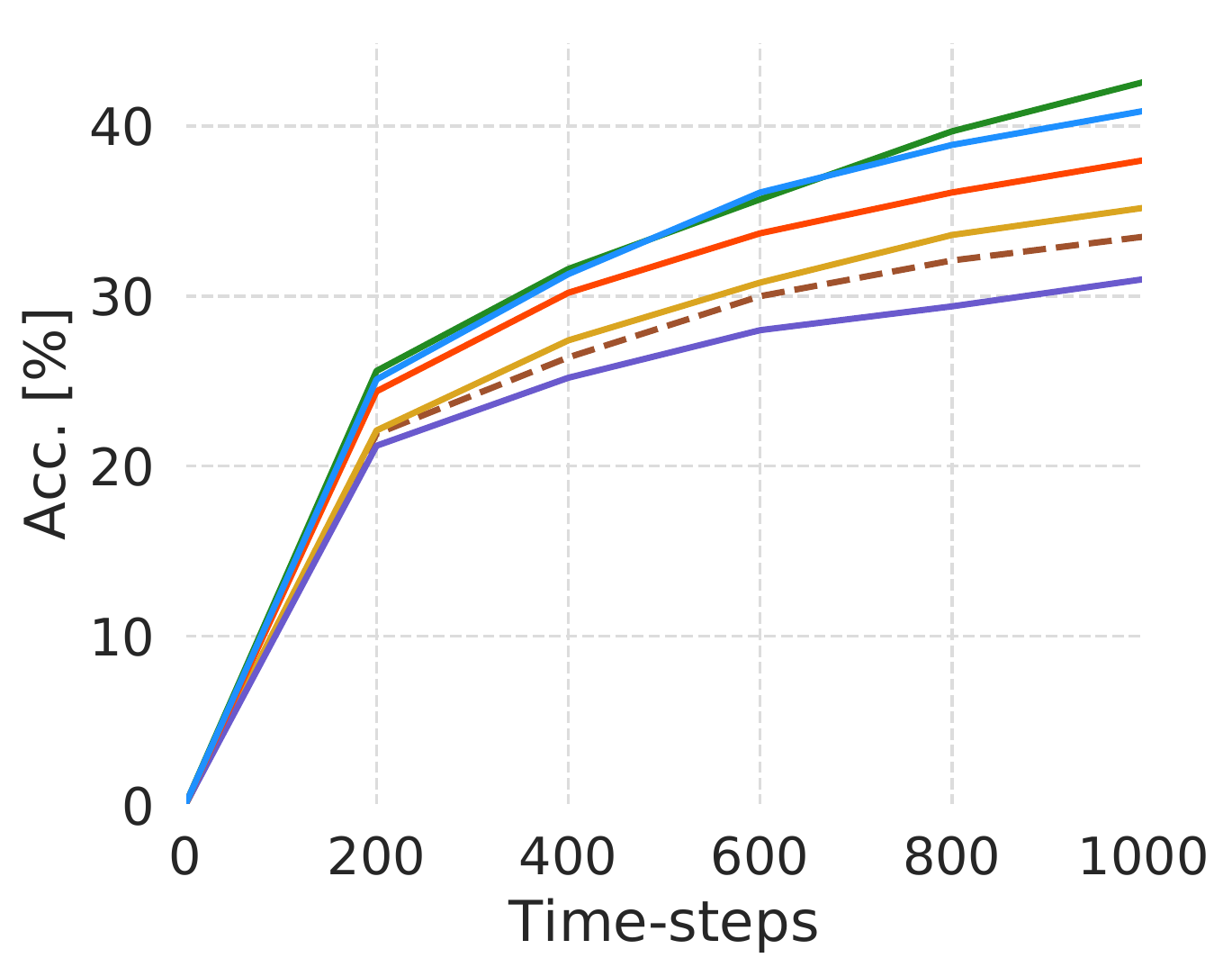} & \includegraphics[width=0.463\linewidth]{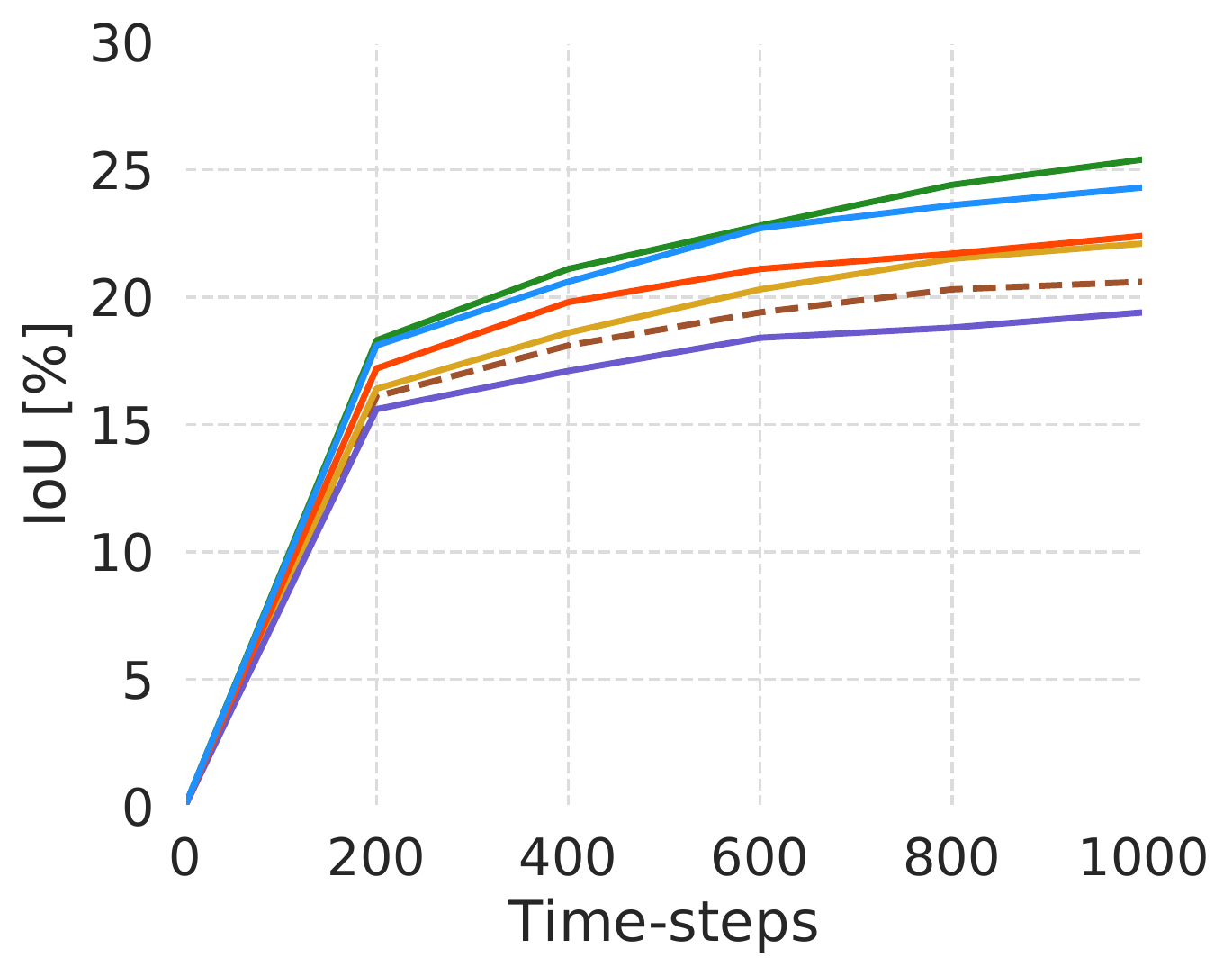} \\
& & \multicolumn{2}{c}{\includegraphics[width=0.93\linewidth]{images/plots_f1/legend_plot_f1.pdf}} \\
\end{tabular}
\caption{Value of accuracy and IoU for the different models at varying time-steps on the MP3D validation set.}
\label{fig:curves_mp3d_val}
\vspace{-0.2cm}
\end{figure}

\begin{figure}[t]
\centering
\scriptsize
\setlength{\tabcolsep}{.2em}
\begin{tabular}{cccc}
& & \textbf{Cumulative $\mathsf{Acc.}$} & \textbf{Cumulative $\mathsf{IoU}$} \\
\rotatebox{90}{\parbox[t]{1.2in}{\hspace*{\fill}\textbf{Estimated Localization}\hspace*{\fill}}} & & 
\includegraphics[width=0.463\linewidth]{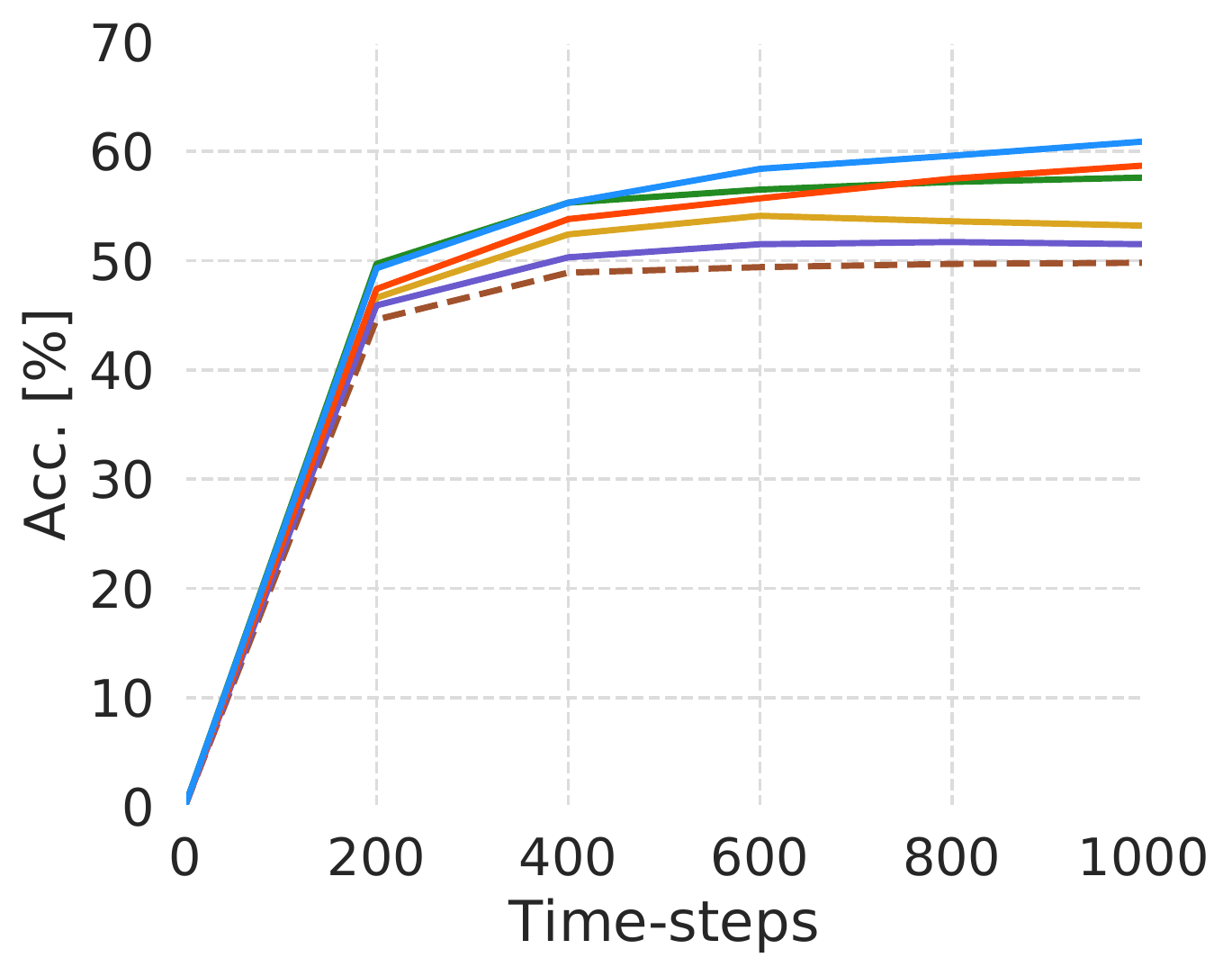} & \includegraphics[width=0.463\linewidth]{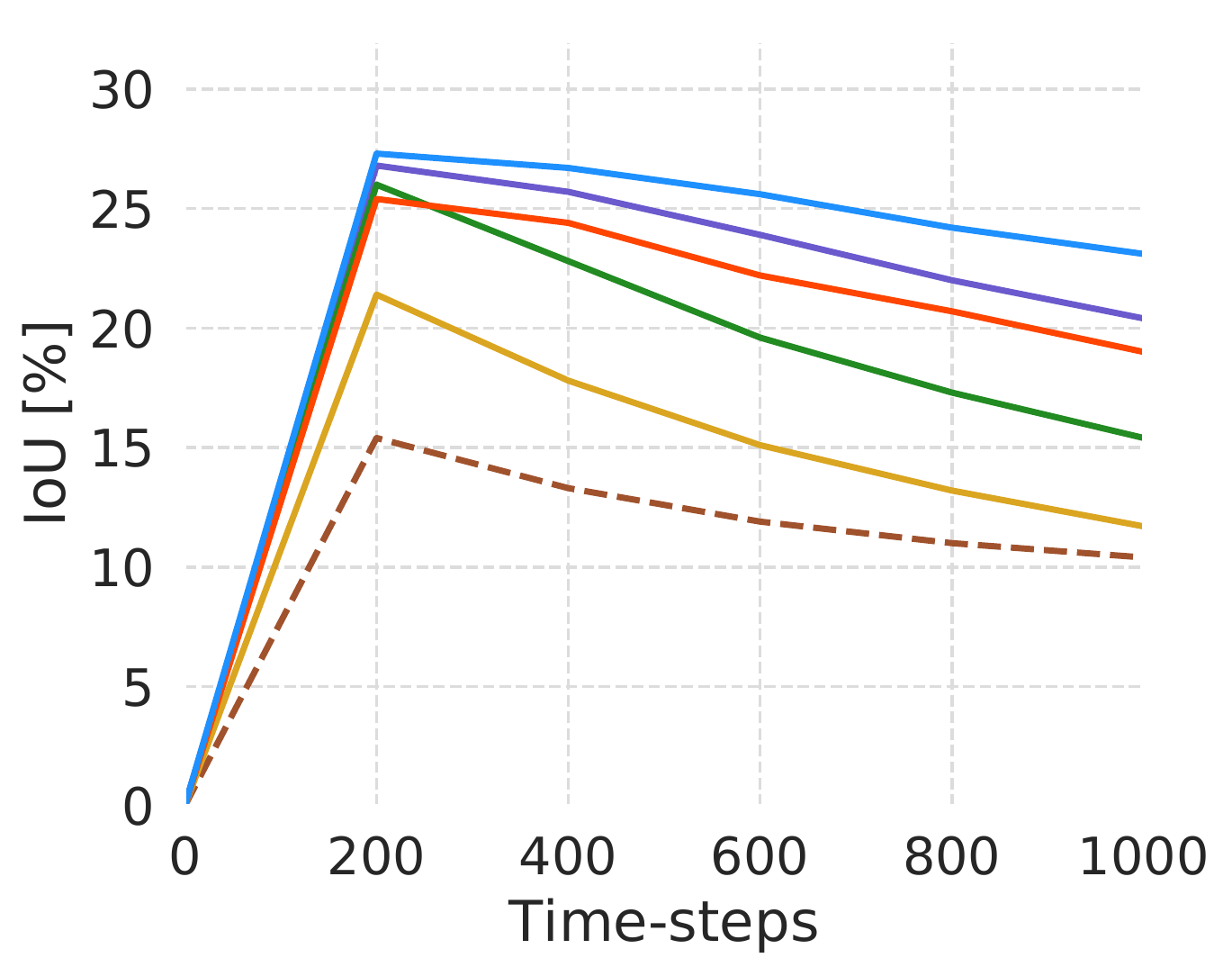} \\ 
\rotatebox{90}{\parbox[t]{1.2in}{\hspace*{\fill}\textbf{Oracle Localization}\hspace*{\fill}}} & &
\includegraphics[width=0.463\linewidth]{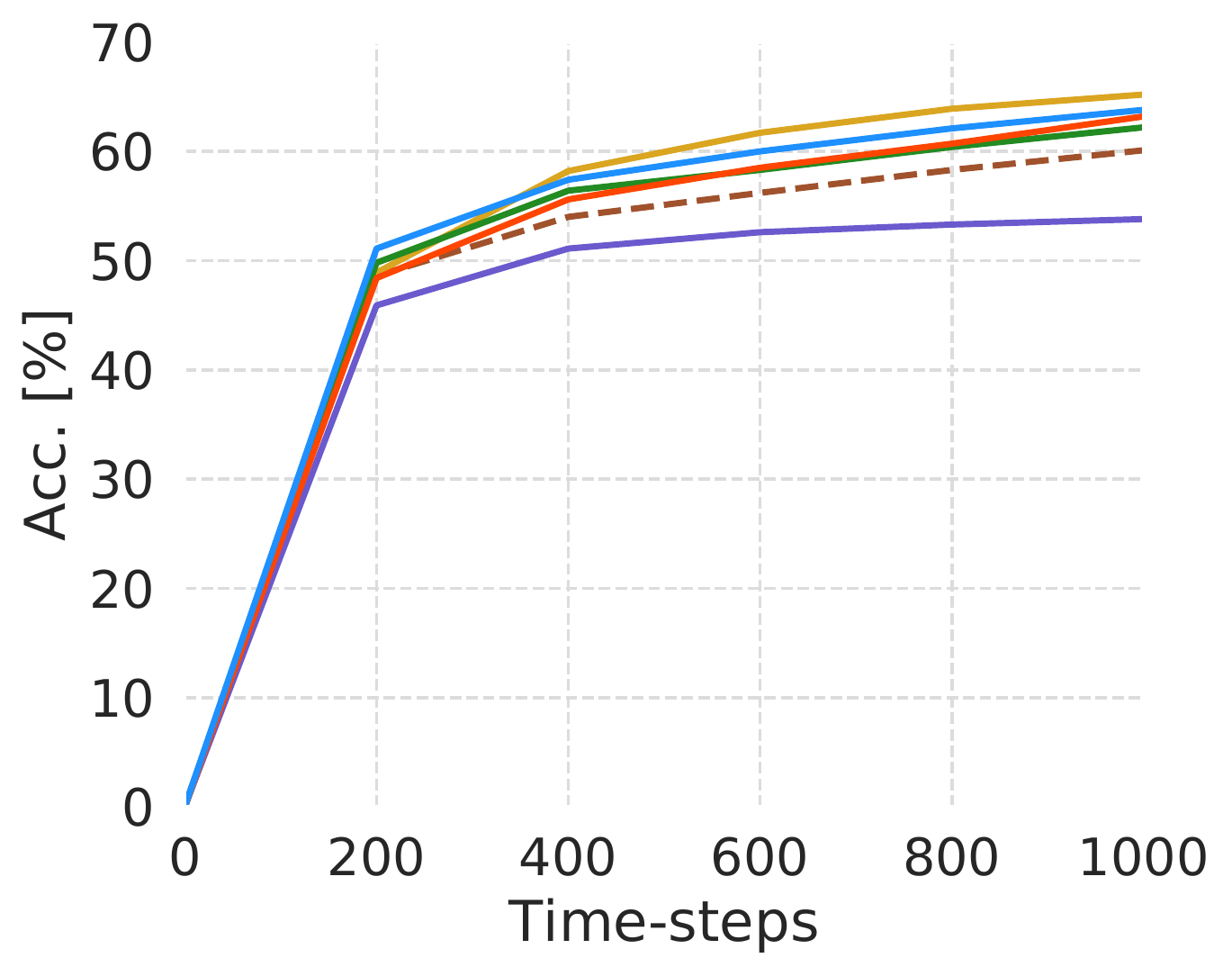} & \includegraphics[width=0.463\linewidth]{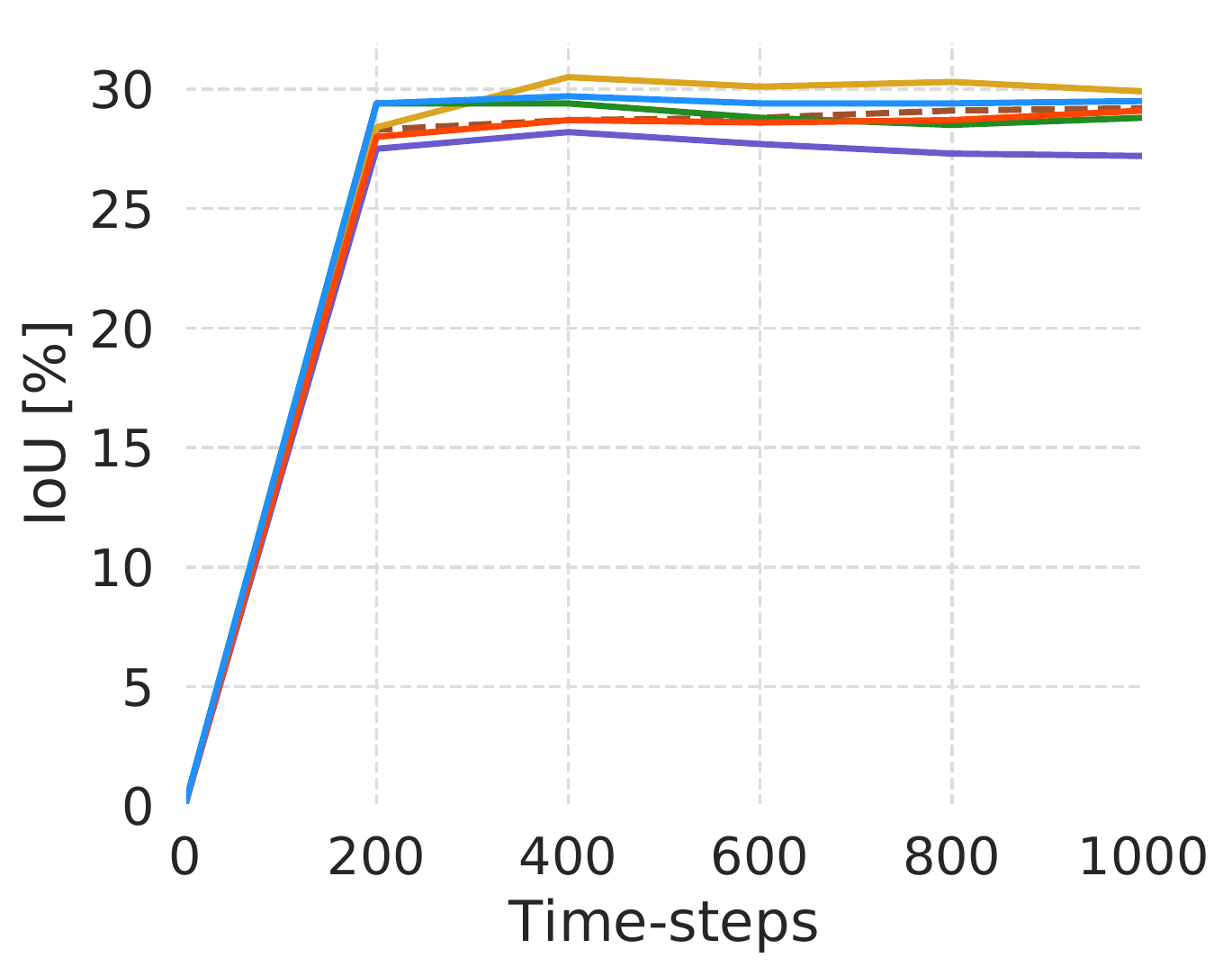} \\
& & \multicolumn{2}{c}{\includegraphics[width=0.93\linewidth]{images/plots_f1/legend_plot_f1.pdf}} \\
\end{tabular}
\caption{Value of accuracy and IoU for the different models at varying time-steps on the Gibson validation set.}
\label{fig:curves_gibson_val}
\vspace{-0.2cm}
\end{figure}

\tit{Analysis at Different Time-Steps}
In Fig.~\ref{fig:curves_mp3d_val} and Fig.~\ref{fig:curves_gibson_val}, we report the plots of different values of $\mathsf{Acc.}$ and $\mathsf{IoU}$ for the MP3D and Gibson validation sets, respectively. In these plots, we show how $\mathsf{Acc.}$ and $\mathsf{IoU}$ vary at different time-steps during the episodes for the various methods.

\begin{figure}[t]
\centering
\scriptsize
\setlength{\tabcolsep}{.2em}
\resizebox{\linewidth}{!}{
\begin{tabular}{cccc}
\textbf{Starting Map} &\textbf{CR} & \textbf{CR+DR} & \textbf{Ground-truth Map} \\
\addlinespace[0.12cm]
\includegraphics[width=0.21\linewidth]{images/supplementary/sample_02_start.png} &
\includegraphics[width=0.21\linewidth]{images/supplementary/sample_02_cr.png} &
\includegraphics[width=0.21\linewidth]{images/supplementary/sample_02_cr_dr.png} & 
\includegraphics[width=0.21\linewidth]{images/supplementary/sample_02_gt.png} \\
\includegraphics[width=0.21\linewidth]{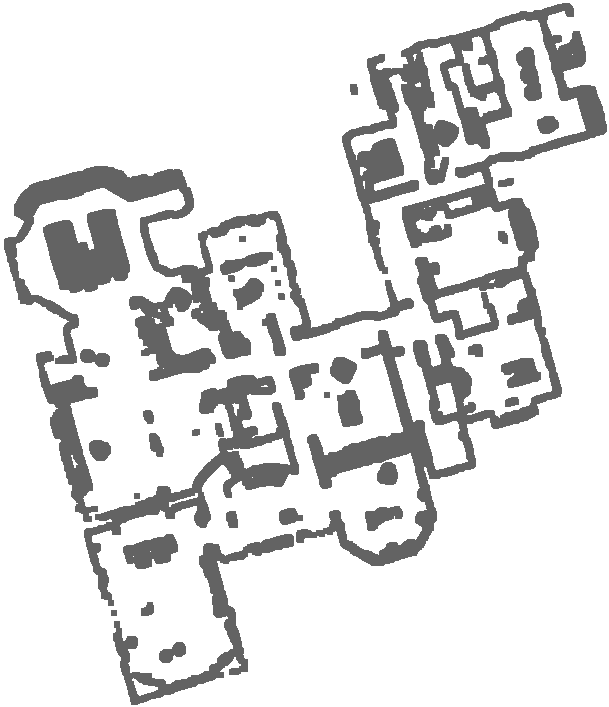} &
\includegraphics[width=0.21\linewidth]{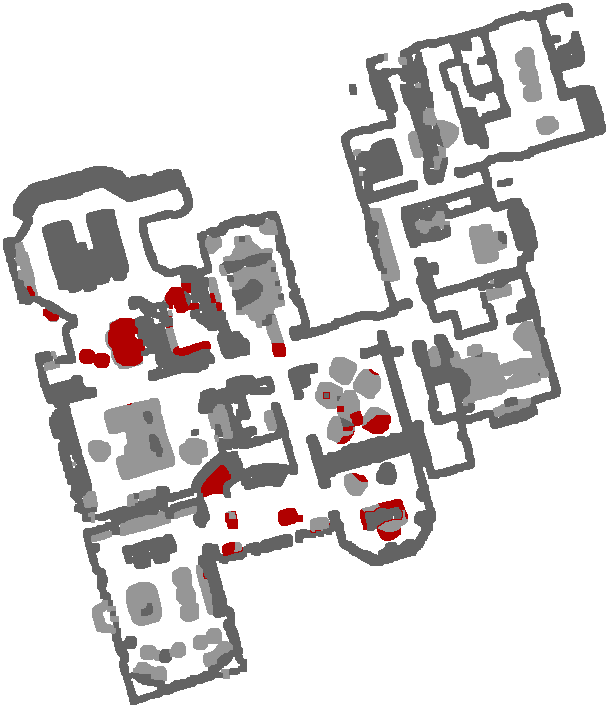} &
\includegraphics[width=0.21\linewidth]{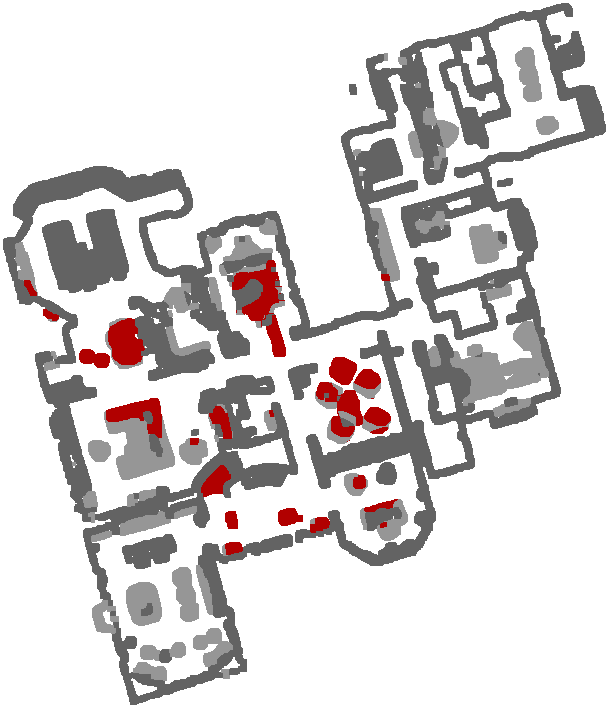} & 
\includegraphics[width=0.21\linewidth]{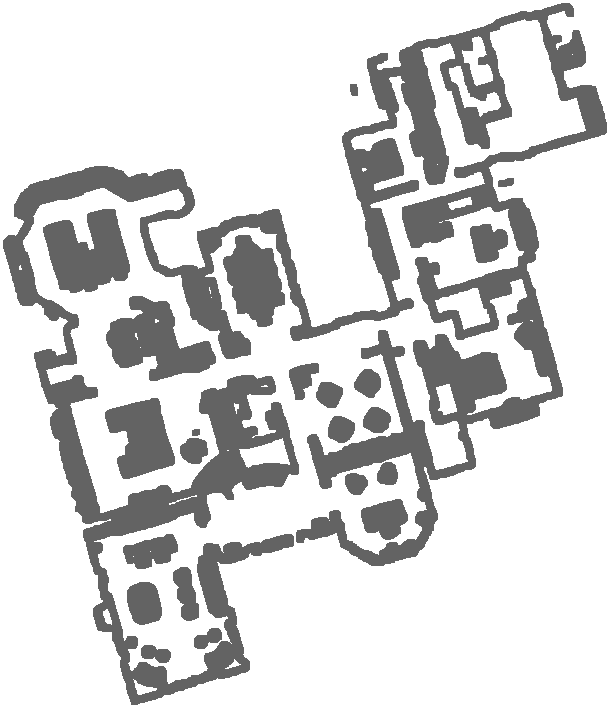} \\
\includegraphics[width=0.21\linewidth]{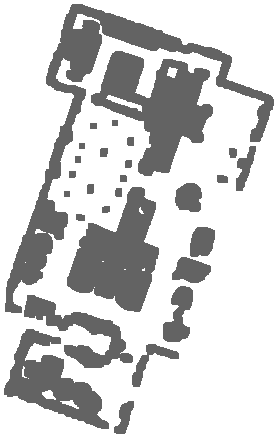} &
\includegraphics[width=0.21\linewidth]{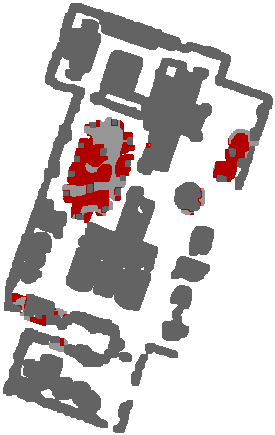} &
\includegraphics[width=0.21\linewidth]{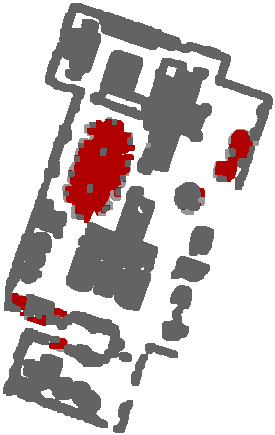} & 
\includegraphics[width=0.21\linewidth]{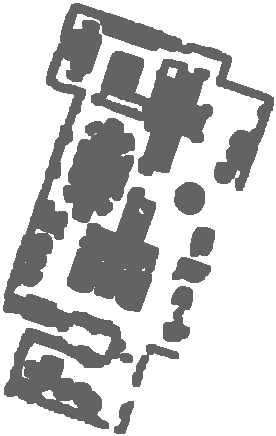} \\
\includegraphics[width=0.21\linewidth]{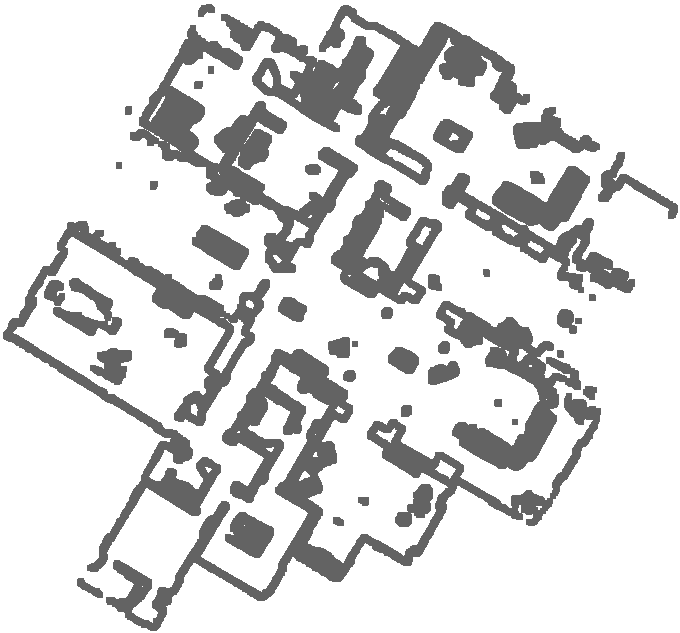} &
\includegraphics[width=0.21\linewidth]{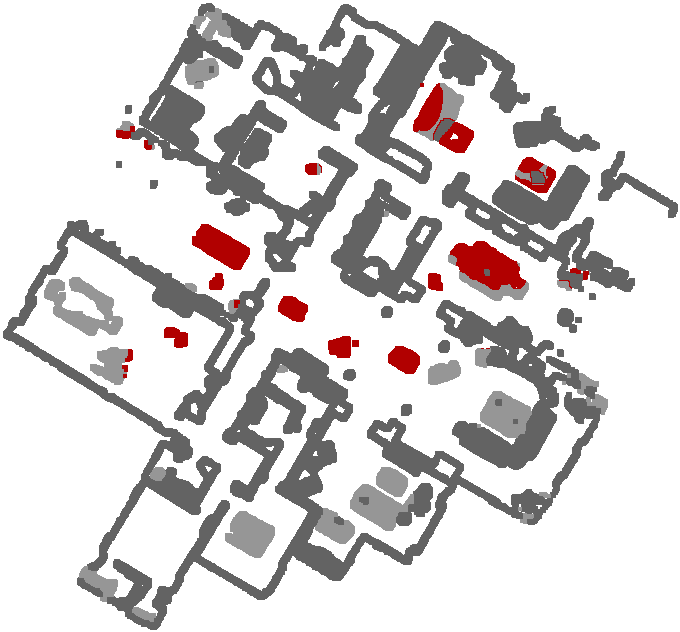} &
\includegraphics[width=0.21\linewidth]{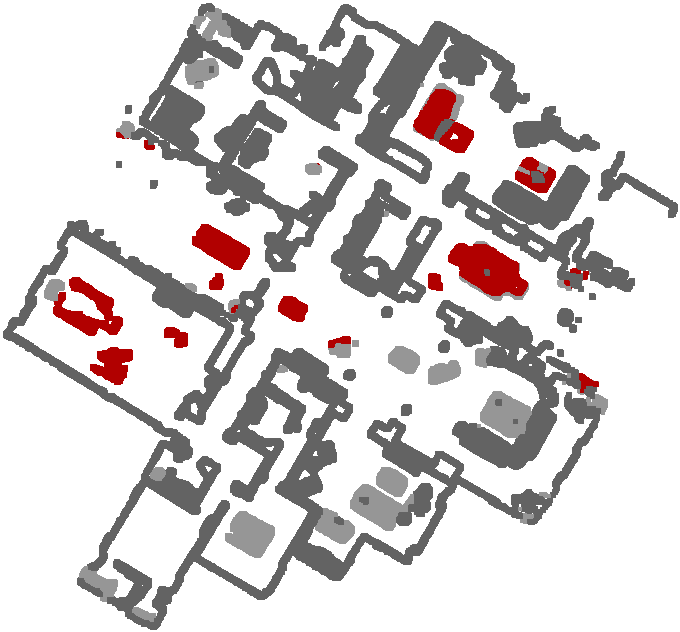} & 
\includegraphics[width=0.21\linewidth]{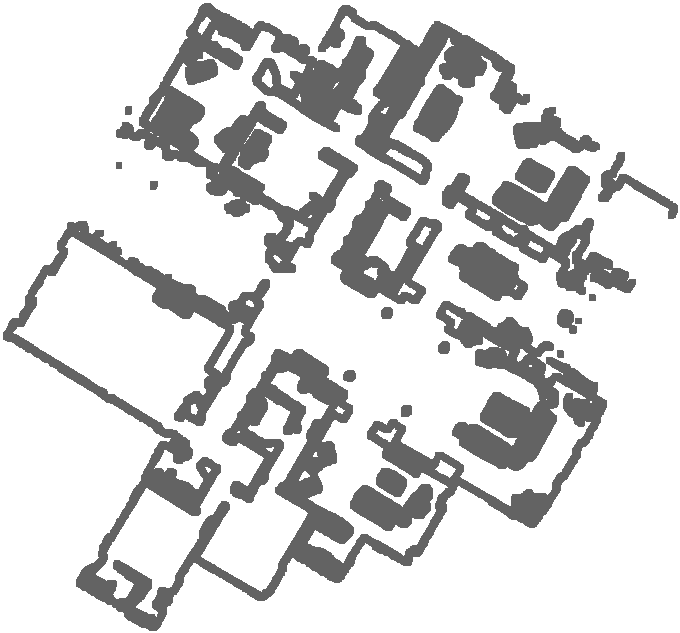} \\
\includegraphics[width=0.21\linewidth]{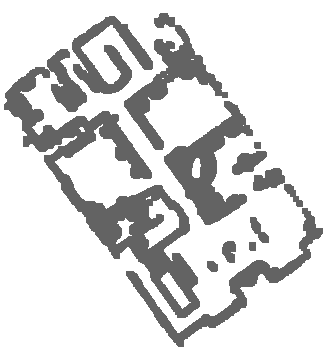} &
\includegraphics[width=0.21\linewidth]{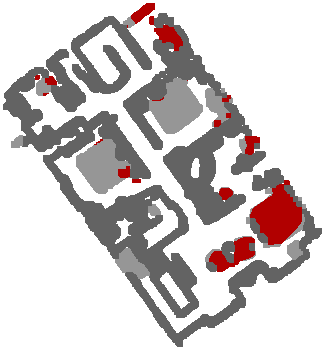} &
\includegraphics[width=0.21\linewidth]{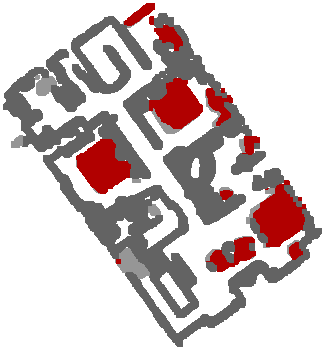} & 
\includegraphics[width=0.21\linewidth]{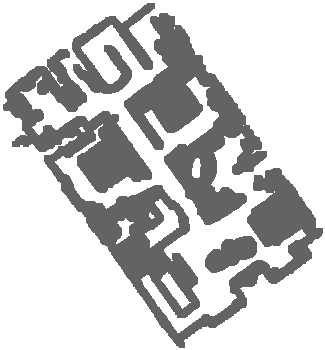} \\
\multicolumn{4}{c}{\includegraphics[width=0.95\linewidth]{images/supplementary/legend_2_f1.pdf}}
\end{tabular}
}
\caption{Qualitative results comparing the performances of the CR and CR+DR agents for different episodes.}
\label{fig:sup_differences}
\vspace{-0.2cm}
\end{figure}

\tit{Qualitative Results}
We include some additional qualitative results for the proposed \textit{Spot the Difference} task in Fig.~\ref{fig:sup_differences}. In particular, we compare the CR agent with the CR+DR counterpart on different episodes with different map size and complexity. For each episode, we report the starting map given to the agent, the reconstructed map collected after exploration, and the ground-truth map of the actual state of the environment. We display the spotted differences in red while keeping the undiscovered elements in light gray. Dark gray denotes unchanged elements of the environment. The proposed CR+DR agent can discover a larger set of differences, thus achieving better results.

\begin{figure}[!t]
\centering
\scriptsize
\setlength{\tabcolsep}{.3em}
\resizebox{\linewidth}{!}{
\begin{tabular}{cc}
\textbf{CR} & \textbf{CR+DR} \\
\addlinespace[0.12cm]
\includegraphics[width=0.49\linewidth]{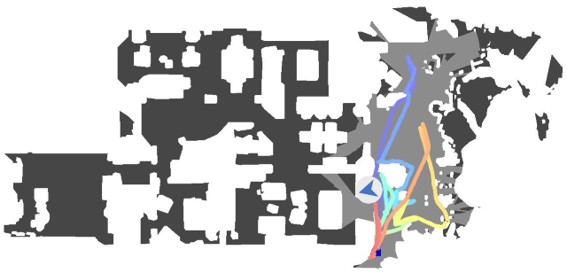} & \includegraphics[width=0.49\linewidth]{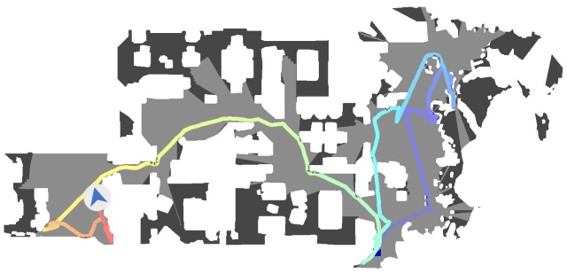} \\ 
\includegraphics[width=0.49\linewidth]{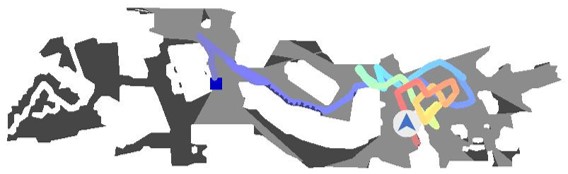} & \includegraphics[width=0.49\linewidth]{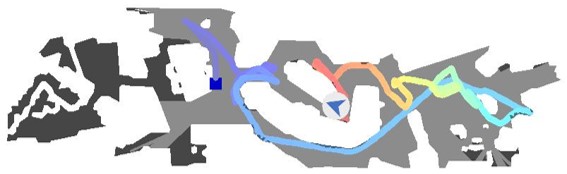} \\ 
\addlinespace[0.2cm]
\multicolumn{2}{c}{\includegraphics[width=0.9\linewidth]{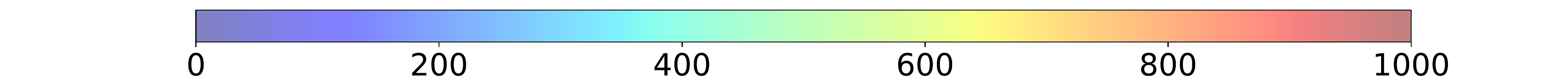}}
\end{tabular}
}
\caption{Exploration trajectories of the CR and CR+DR agents on sample MP3D test episodes.}
\label{fig:navigation}
\end{figure}

Moreover, we report sample exploration trajectories for the CR and the CR+DR agents with estimated localization in Fig.~\ref{fig:navigation}. These confirm the competitive exploration capabilities of our proposed agent.

\subsection{Discussion and Future Directions}
We present a method that exploits outdated information about the current environment to improve the exploration capabilities of the agent. However, the focus of this work is on pure occupation, ignoring semantic information. For future work, we expect to include semantic reasoning into the agent's pipeline. We assume that additional information could boost the performance.
With the proposed dataset, we enable a series of possible embodied tasks that imply dynamic environments and incorporate available past knowledge.

\begin{table*}[t]
\begin{minipage}[t]{.45\textwidth}
\vspace{1.8cm}
\centering
\setlength{\tabcolsep}{.35em}
\footnotesize
\caption{MP3D semantic categories per channel index.}
\label{tab:semantic_classes_mp3d}
\resizebox{0.8\linewidth}{!}{
\begin{tabular}{llc}
\toprule
\multicolumn{3}{c}{\textbf{MP3D}} \\
\midrule
\textbf{Index} & \textbf{Category} & \textbf{Action} \\
\midrule
0 & Void & No Operation \\
1 & Wall & No Operation \\
2 & Floor & No Operation \\
3 & Chair & Displacement \\
4 & Door & No Operation \\
5 & Table & Displacement \\
6 & Picture & No Operation \\
7 & Cabinet & Removal \\
8 & Cushion & Overlap Removal \\
9 & Window & No Operation \\
10 & Sofa & Displacement \\
11 & Bed & Displacement \\
12 & Curtain & No Operation \\
13 & Chest of Drawers & Displacement \\
14 & Plant & Displacement \\
15 & Sink & Empty \\
16 & Stairs & No Operation \\
17 & Ceiling & No Operation \\
18 & Toilet & Removal \\
19 & Stool & Displacement \\
20 & Towel & Overlap Removal \\
21 & Mirror & No Operation \\
22 & TV Monitor & Removal \\
23 & Shower & Removal \\
24 & Column & No Operation \\
25 & Bathtub & Removal \\
26 & Counter & Removal \\
27 & Fireplace & No Operation \\
28 & Lighting & No Operation \\
29 & Beam & No Operation \\
30 & Railing & No Operation \\
31 & Shelving & Removal \\
32 & Blinds & No Operation \\
33 & Gym Equipment & Displacement \\
34 & Seating & Removal \\
35 & Board Panel & No Operation \\
36 & Furniture & Displacement \\
37 & Appliances & Removal \\
38 & Clothes & Overlap Removal \\
39 & Objects & Overlap Removal \\
40 & Misc & Overlap Removal \\
41 & Unlabeled & No Operation \\
\bottomrule
\end{tabular}
}
\end{minipage}
\begin{minipage}[t]{.45\textwidth}
\centering
\setlength{\tabcolsep}{.35em}
\footnotesize
\caption{Gibson semantic categories per channel index.}
\label{tab:semantic_classes_gibson}
\resizebox{0.8\linewidth}{!}{
\begin{tabular}{llc}
\toprule
\multicolumn{3}{c}{\textbf{Gibson}} \\
\midrule
\textbf{Index} & \textbf{Category} & \textbf{Action} \\
\midrule
0 & Chair & Displacement \\
1 & Couch & Displacement \\
2 & Potted Plant & Removal \\
3 & Bed & Displacement \\
4 & Toilet & Removal \\
5 & TV & Removal \\
6 & Dining Table & Displacement \\
7 & Oven & Removal \\
8 & Sink & Removal \\
9 & Refrigerator & Removal \\
10 & Book & Overlap Removal \\
11 & Clock & Removal \\
12 & Vase & Removal \\
13 & Cup & Overlap Removal \\
14 & Bottle & Overlap Removal \\
15 & Bench & Removal \\
16 & Appliances & Removal \\
17 & Objects & Overlap Removal \\
18 & Misc & Overlap Removal \\
19 & Void & No Operation \\
\bottomrule
\end{tabular}
}
\vspace{.5cm}
\centering
\footnotesize
\setlength{\tabcolsep}{.35em}
\caption{List of hyperparameters.}
\label{tab:hyperparams}
\resizebox{0.8\linewidth}{!}{
\begin{tabular}{lll}
\toprule
\multicolumn{3}{c}{\textbf{Hyperparameters}} \\
\midrule
\textbf{Module} & \textbf{Name} &  \textbf{Value} \\
\midrule

Global Policy   & map size (G)      & $240$ \\
                & lr                & $2.5\times10^{-4}$ \\
                & max\_grad\_norm   & $0.5$ \\
\addlinespace[0.2cm]
Local Policy    & forward step      & $0.25m$ \\
                & turn angle        & $10\degree$ \\
                & hidden size       & $256$ \\
                & lr                & $2.5\times10^{-4}$ \\
\addlinespace[0.2cm]
Mapper          & batch size        & $32$ \\
                & map scale         & $0.05cm^2$ \\
                & map size          & $101$ \\
                & lr                & $10^{-3}$ \\
                & momentum          & $0.9$ \\
                & max grad norm     & $0.5$ \\
\addlinespace[0.2cm]
PPO             & clip param        & $0.2$ \\
                & entropy coef      & $10^{-3}$ \\
                & eps               & $10^{-5}$ \\
                & gamma             & $0.99$ \\
                & n.~mini-batches   & $4$ \\
                & n.~epochs         & $4$ \\
                & tau               & $0.95$ \\
                & using gae         & True \\
                & value loss coef   & $0.5$ \\
\bottomrule
\end{tabular}
}
\end{minipage}
\end{table*}

\begin{table*}[t]
\begin{minipage}[t]{.45\textwidth}
\centering
\caption{MP3D train scans and floors.} 
\label{tab:sup_mp3dtrain}
\setlength{\tabcolsep}{.35em}
\footnotesize
\resizebox{0.8\linewidth}{!}{
\begin{tabular}{lccc}
\toprule
\multicolumn{4}{c}{\textbf{MP3D Train}} \\
\midrule
\textbf{Scan} & & \textbf{Floors} & \textbf{\# Episodes} \\
\midrule
HxpKQynjfin & & 0 & 81967 \\
gTV8FGcVJC9 & & 0,1,2,3,4,6,10,11 & 77186 \\
29hnd4uzFmX & & 0,1,2,3 & 81967 \\
5LpN3gDmAk7 & & 0,1,2,3 & 81885 \\
SN83YJsR3w2 & & 0,1,2,3,7,8,10,12 & 81438 \\
VzqfbhrpDEA & & 0,1,3,6 & 81641 \\
D7N2EKCX4Sj & & 0,1,2,3,5,6 & 81830 \\
5q7pvUzZiYa & & 0,1,2,3,4 & 81967 \\
ac26ZMwG7aT & & 0,1 & 81967 \\
r47D5H71a5s & & 0,1 & 81965 \\
Pm6F8kyY3z2 & & 0 & 81967 \\
8WUmhLawc2A & & 0,1,2 & 81967 \\
82sE5b5pLXE & & 0,1,2 & 80682 \\
mJXqzFtmKg4 & & 0,1,2 & 81967 \\
i5noydFURQK & & 0,1 & 81120 \\
V2XKFyX4ASd & & 0,1,2,3,4,5,7 & 81129 \\
759xd9YjKW5 & & 0,1,2,3 & 81913 \\
r1Q1Z4BcV1o & & 0 & 81812 \\
S9hNv5qa7GM & & 0,1 & 81967 \\
1LXtFkjw3qL & & 0,1,2,3,4,5,6 & 81967 \\
PuKPg4mmafe & & 0 & 81940 \\
EDJbREhghzL & & 0,1,3 & 64755 \\
ur6pFq6Qu1A & & 0,1 & 81967 \\
B6ByNegPMKs & & 0 & 81951 \\
b8cTxDM8gDG & & 0,1,2,8,11 & 73307 \\
17DRP5sb8fy & & 0 & 81967 \\
YmJkqBEsHnH & & 0 & 80780 \\
ULsKaCPVFJR & & 0,1,2 & 81967 \\
XcA2TqTSSAj & & 0,2,3,5,6,8,9,11,12 & 60679 \\
sKLMLpTHeUy & & 0,1,2,4 & 79736 \\
ZMojNkEp431 & & 0,1,2 & 81967 \\
e9zR4mvMWw7 & & 0,1,2 & 80193 \\
JeFG25nYj2p & & 0,1 & 81967 \\
uNb9QFRL6hY & & 1,4,5,6 & 59613 \\
p5wJjkQkbXX & & 0,1,2,3 & 81967 \\
Vvot9Ly1tCj & & 0,3 & 78115 \\
E9uDoFAP3SH & & 0,1,5,6 & 81914 \\
qoiz87JEwZ2 & & 0,1,2,3 & 81967 \\
VFuaQ6m2Qom & & 0,1,2,4,5,6 & 81758 \\
VLzqgDo317F & & 0,1,2 & 81396 \\
kEZ7cmS4wCh & & 0,1,2,3,7 & 69135 \\
7y3sRwLe3Va & & 0,1,2,5 & 81386 \\
VVfe2KiqLaN & & 0,1,2 & 81967 \\
2n8kARJN3HM & & 0,1,2,4 & 81076 \\
PX4nDJXEHrG & & 0,1,2,3,4,5 & 79151 \\
Uxmj2M2itWa & & 0,1,3,4 & 49942 \\
pRbA3pwrgk9 & & 0,2,3,7,9,11 & 53295 \\
cV4RVeZvu5T & & 0,1,2,3 & 81038 \\
sT4fr6TAbpF & & 0 & 81625 \\
GdvgFV5R1Z5 & & 0 & 81967 \\
JF19kD82Mey & & 0,1,2 & 81927 \\
JmbYfDe2QKZ & & 0,1 & 81489 \\
s8pcmisQ38h & & 0,1,2 & 80428 \\
1pXnuDYAj8r & & 0,1,2,5 & 81901 \\
jh4fc5c5qoQ & & 0,1,2 & 81967 \\
vyrNrziPKCB & & 0,1,3,4,7 & 81388 \\
aayBHfsNo7d & & 0,1,2 & 81693 \\
rPc6DW4iMge & & 0,1,3,4 & 80296 \\
\midrule
\textbf{Total:} 58 & & 207 & 4581881 \\
\bottomrule
\end{tabular}
}
\end{minipage}
\begin{minipage}[t]{.4\textwidth}
\vspace{0.65cm}
\centering
\caption{MP3D validation scans and floors, with relative number of episodes for \textit{Spot the Difference}.}
\label{tab:sup_mp3dval}
\setlength{\tabcolsep}{.35em}
\footnotesize
\resizebox{0.7\linewidth}{!}{
\begin{tabular}{lccc}
\toprule
\multicolumn{4}{c}{\textbf{MP3D Validation}} \\
\midrule
\textbf{Scan} & & \textbf{Floors} & \textbf{\# Episodes} \\
\midrule
2azQ1b91cZZ & & 0,1 & 40 \\
8194nk5LbLH & & 0 & 40 \\
EU6Fwq7SyZv & & 0 & 30 \\
QUCTc6BB5sX & & 1 & 20 \\
TbHJrupSAjP & & 0,1,2 & 30 \\
Z6MFQCViBuw & & 0 & 40 \\
oLBMNvg9in8 & & 0,1,2,3 & 50 \\
x8F5xyUWy9e & & 0,1 & 30 \\
zsNo4HB9uLZ & & 0 & 40 \\
\midrule
\textbf{Total:} 9 & & 16 & 320 \\
\bottomrule
\end{tabular}
}
\vspace{1cm}
\centering
\caption{MP3D test scans and floors, with relative number of episodes for \textit{Spot the Difference}.}
\label{tab:sup_mp3dtest}
\setlength{\tabcolsep}{.35em}
\footnotesize
\resizebox{0.7\linewidth}{!}{
\begin{tabular}{lccc}
\toprule
\multicolumn{4}{c}{\textbf{MP3D Test}} \\
\midrule
\textbf{Scan} & & \textbf{Floors} & \textbf{\# Episodes} \\
\midrule
2t7WUuJeko7 & & 0 & 50 \\
5ZKStnWn8Zo & & 0,1 & 50 \\
RPmz2sHmrrY & & 0 & 50 \\
UwV83HsGsw3 & & 0,1,2,3 & 50 \\
WYY7iVyf5p8 & & 0,2 & 30 \\
YFuZgdQ5vWj & & 1 & 10 \\
YVUC4YcDtcY & & 0 & 50 \\
fzynW3qQPVF & & 0,1 & 50 \\
jtcxE69GiFV & & 0,1 & 40 \\
pa4otMbVnkk & & 0,1 & 50 \\
q9vSo1VnCiC & & 0 & 50 \\
rqfALeAoiTq & & 0,2 & 20 \\
wc2JMjhGNzB & & 0,1 & 50 \\
yqstnuAEVhm & & 0,1,2 & 60 \\
\midrule
\textbf{Total:} 14 & & 26 & 610 \\
\bottomrule
\end{tabular}
}
\vspace{1cm}
\centering
\caption{Gibson validation scans and floors, with relative number of episodes for \textit{Spot the Difference}.}
\label{tab:sup_gibsonval}
\footnotesize
\setlength{\tabcolsep}{.35em}
\resizebox{0.7\linewidth}{!}{
\begin{tabular}{lccc}
\toprule
\multicolumn{4}{c}{\textbf{Gibson Validation}} \\
\midrule
\textbf{Scan} & & \textbf{Floors} & \textbf{\# Episodes} \\
\midrule
Wiconisco & & 1,2 & 90 \\
Corozal & & 0,2,4 & 90 \\
Collierville & & 0,1,2 & 80 \\
Markleeville & & 0,1 & 90 \\
Darden & & 0,1,2 & 100 \\
\midrule
\textbf{Total:} 5 & & 13 & 450 \\
\bottomrule
\end{tabular}
}
\end{minipage}
\end{table*}

\begin{figure*}[t]
\centering
\scriptsize
\setlength{\tabcolsep}{.2em}
\begin{tabular}{cccccc}
\textbf{Original Map} & & \textbf{Manipulated Map 1} & \textbf{Difference Map 1}  & \textbf{Manipulated Map 2}  & \textbf{Difference Map 2} \\
\addlinespace[0.12cm]
\includegraphics[width=0.182\linewidth]{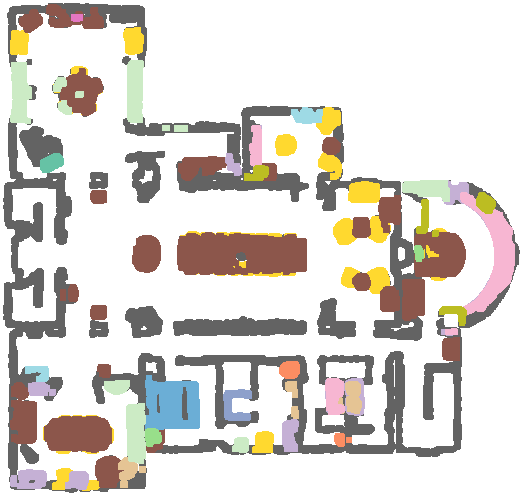} & & 
\includegraphics[width=0.192\linewidth]{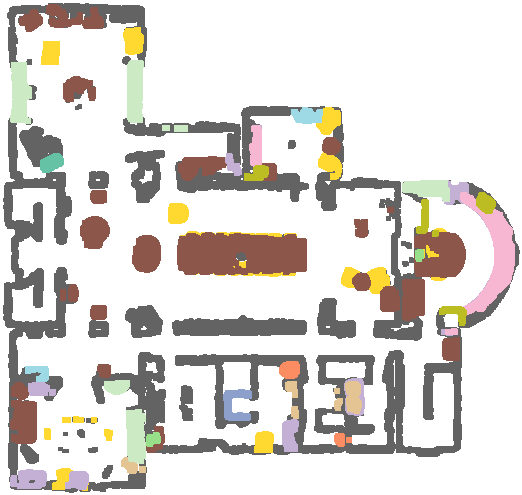} &
\includegraphics[width=0.192\linewidth]{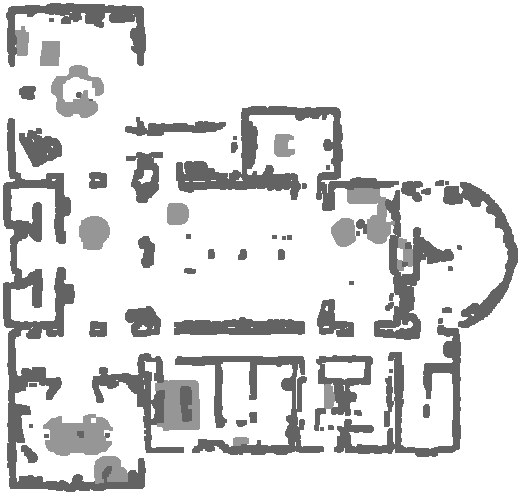} &
\includegraphics[width=0.192\linewidth]{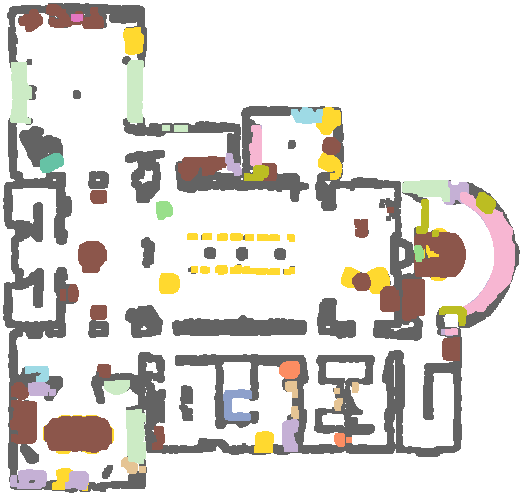} &
\includegraphics[width=0.192\linewidth]{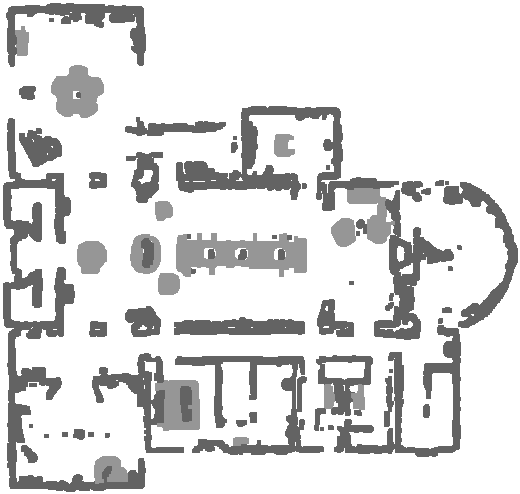} \\
\addlinespace[0.05cm]
\includegraphics[width=0.192\linewidth]{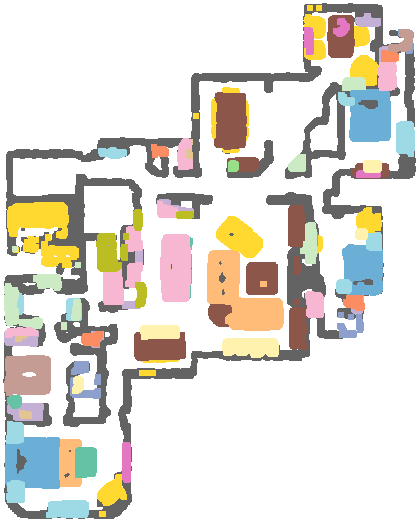} & & 
\includegraphics[width=0.192\linewidth]{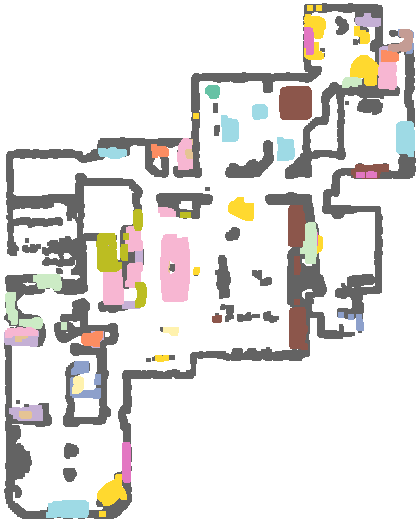} &
\includegraphics[width=0.192\linewidth]{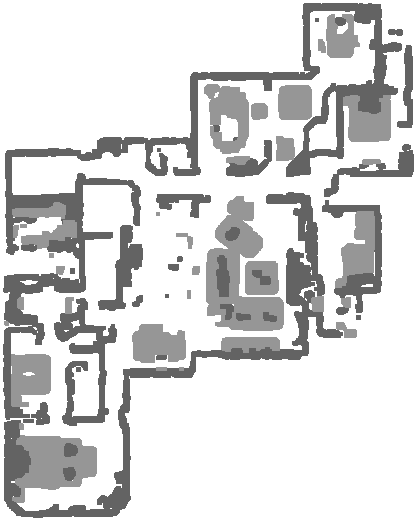} &
\includegraphics[width=0.192\linewidth]{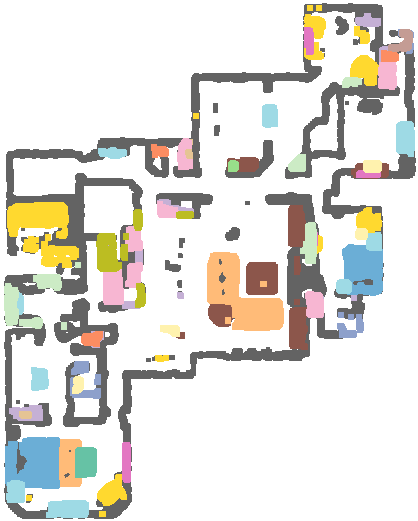} &
\includegraphics[width=0.192\linewidth]{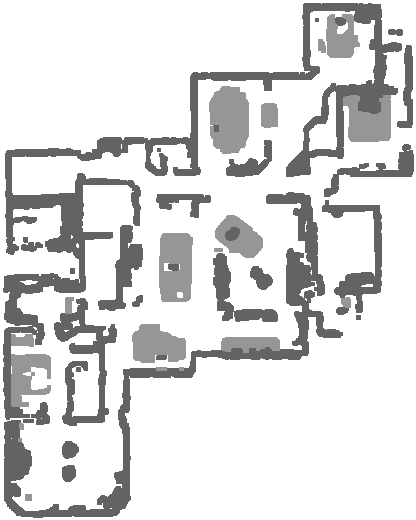} \\
\addlinespace[0.05cm]
\includegraphics[width=0.192\linewidth]{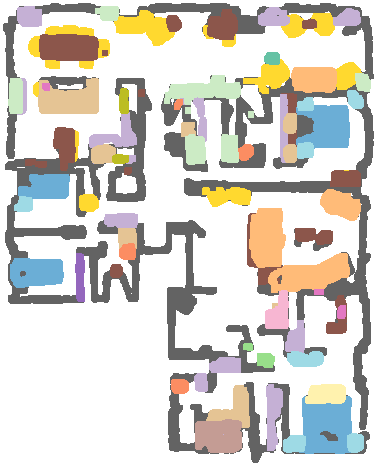} & & 
\includegraphics[width=0.192\linewidth]{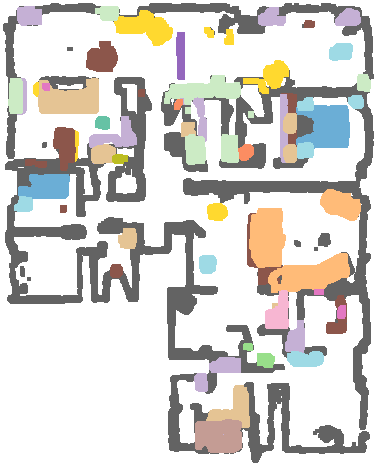} &
\includegraphics[width=0.192\linewidth]{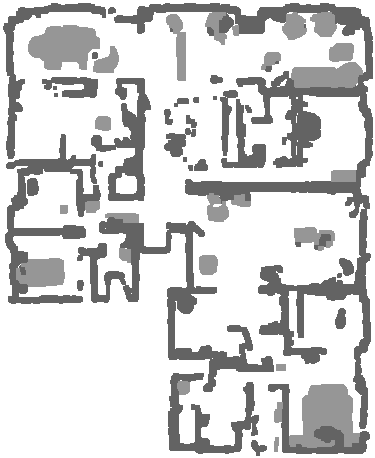} &
\includegraphics[width=0.192\linewidth]{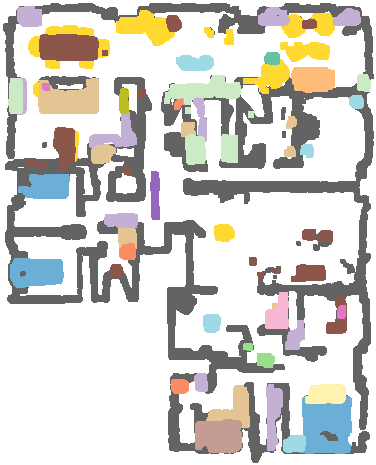} &
\includegraphics[width=0.192\linewidth]{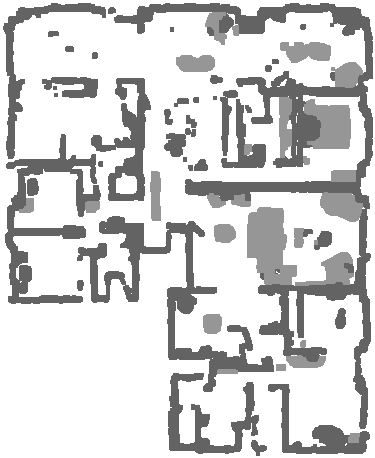} \\
\addlinespace[0.05cm]
\includegraphics[width=0.192\linewidth]{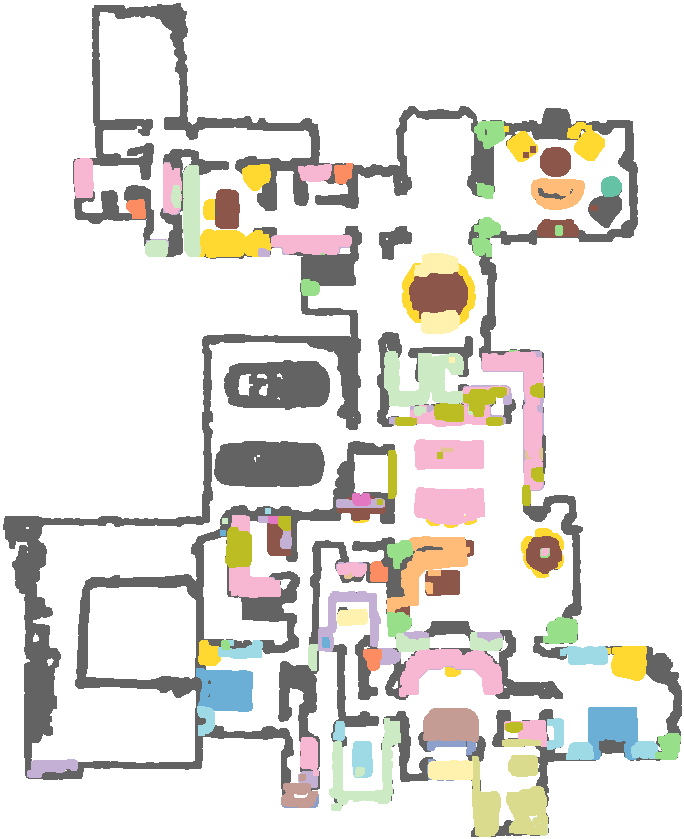} & & 
\includegraphics[width=0.192\linewidth]{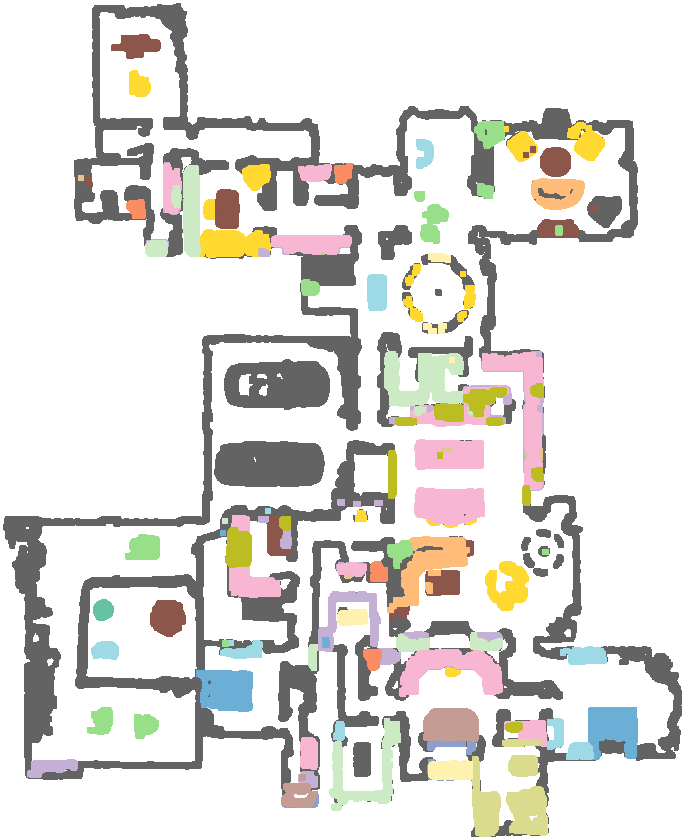} &
\includegraphics[width=0.192\linewidth]{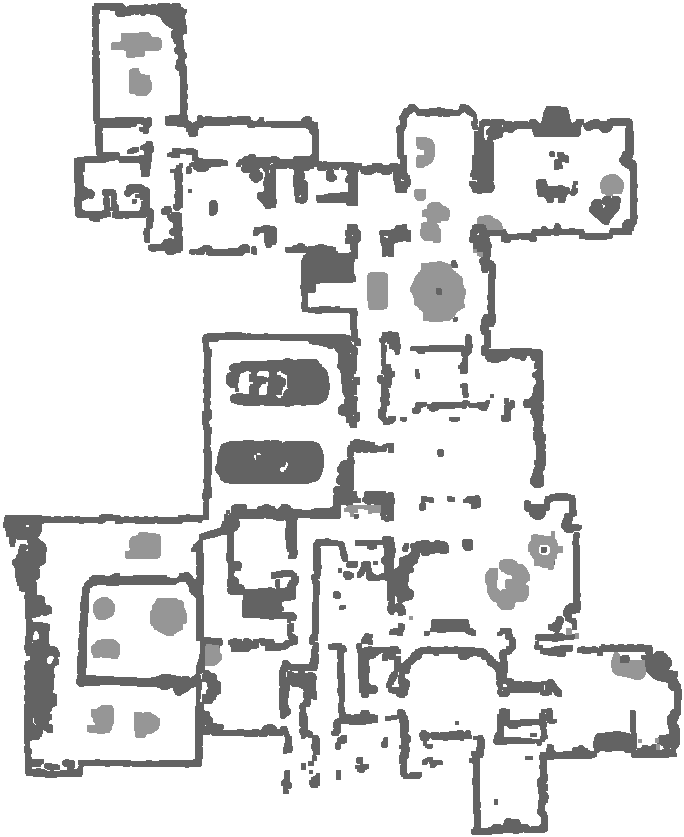} &
\includegraphics[width=0.192\linewidth]{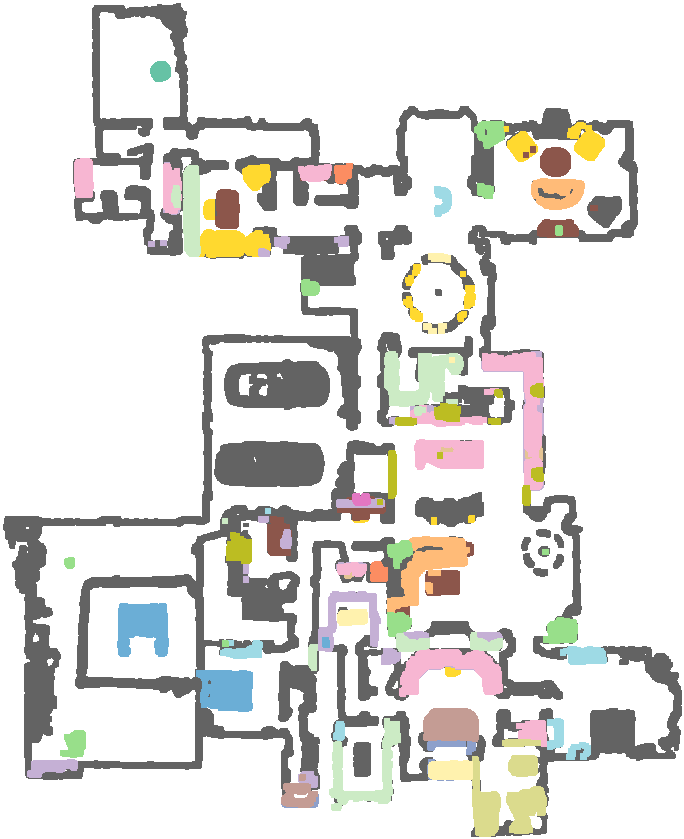} &
\includegraphics[width=0.192\linewidth]{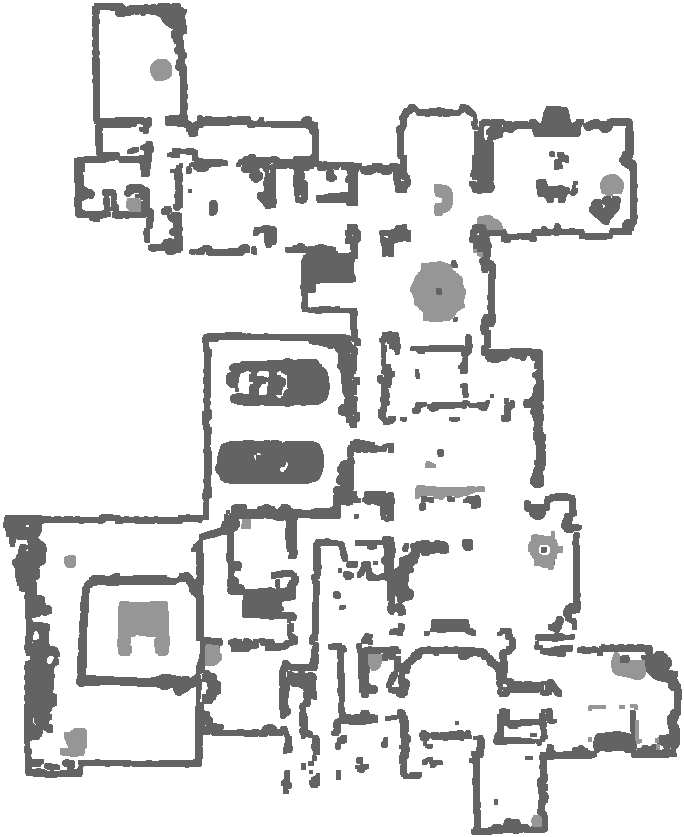} \\
\addlinespace[0.05cm]
\includegraphics[width=0.192\linewidth]{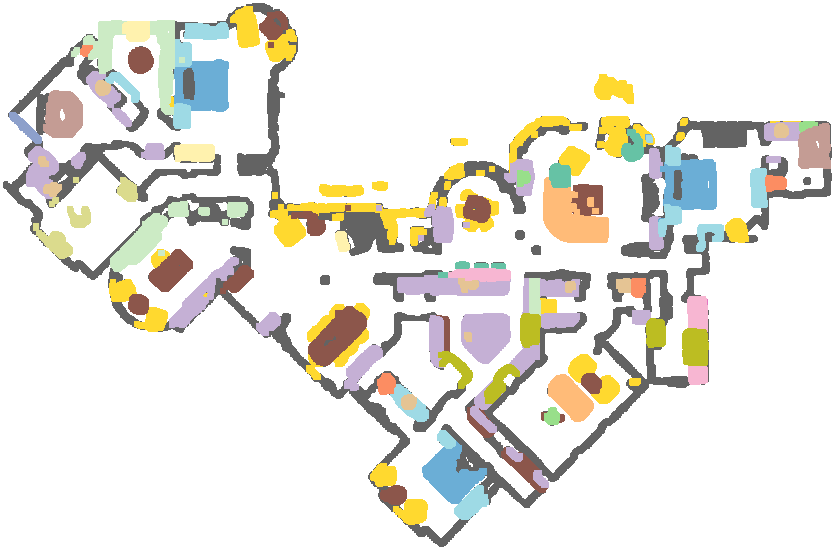} & & 
\includegraphics[width=0.192\linewidth]{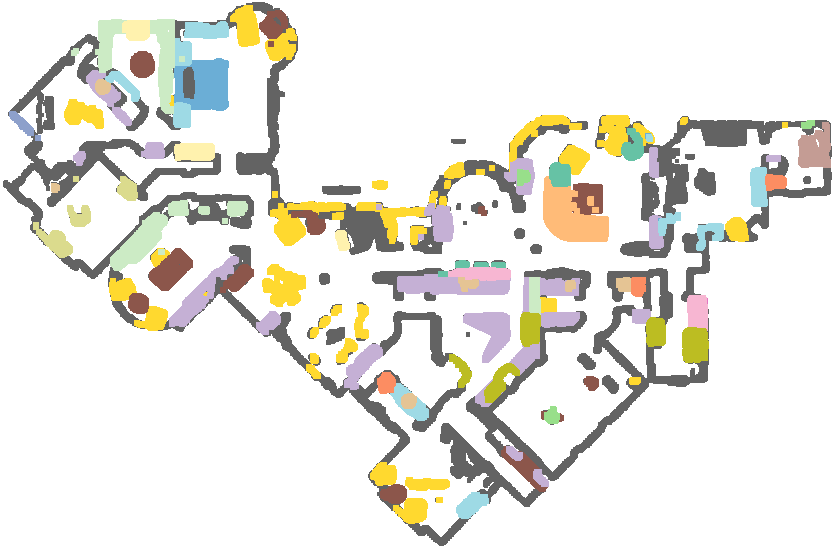} &
\includegraphics[width=0.192\linewidth]{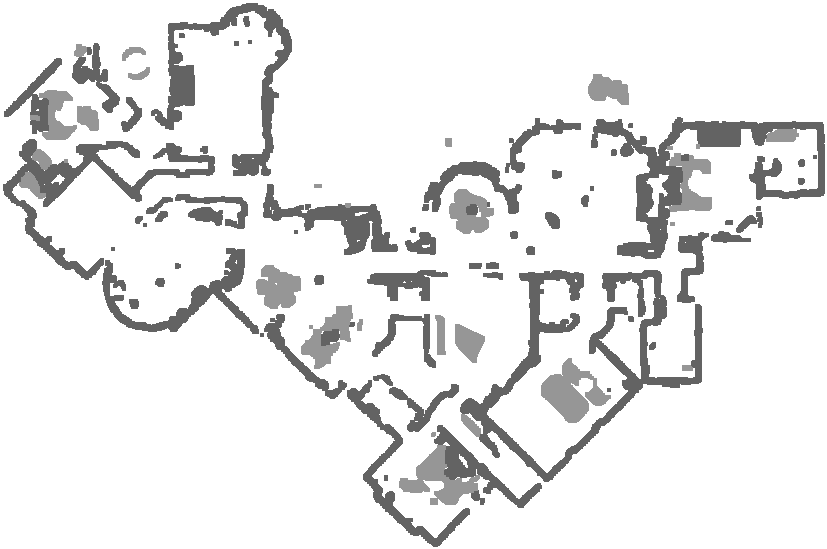} &
\includegraphics[width=0.192\linewidth]{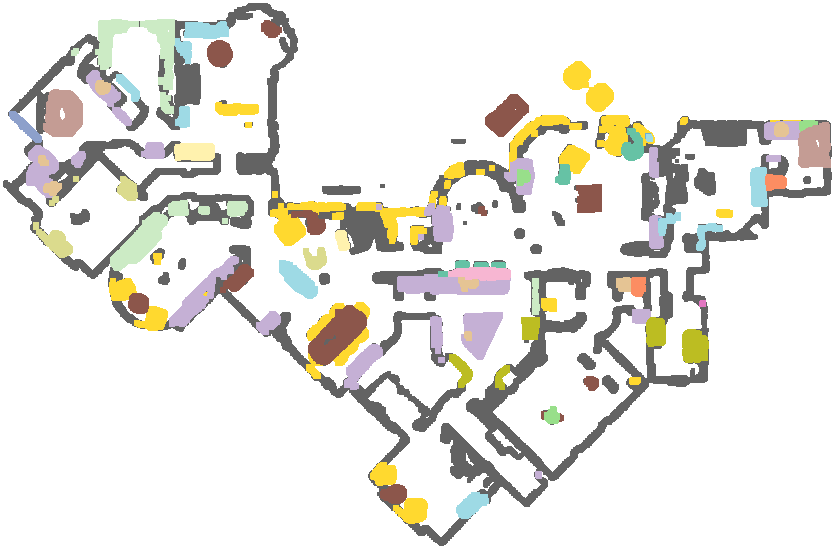} &
\includegraphics[width=0.192\linewidth]{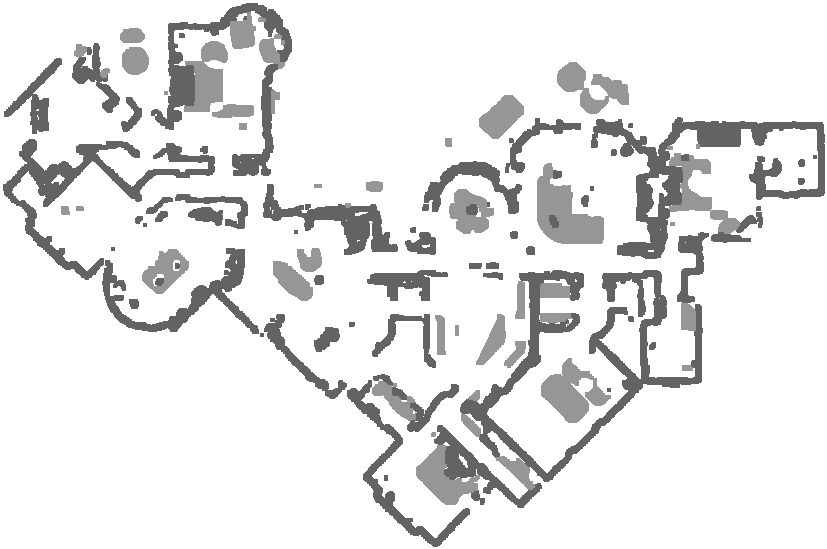} \\
\addlinespace[0.12cm]
\multicolumn{6}{c}{\includegraphics[width=0.95\linewidth]{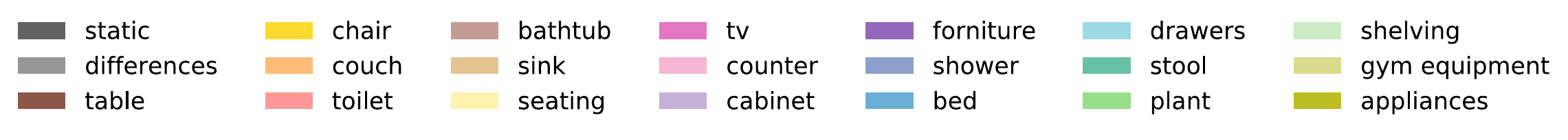}} \\
\end{tabular}
\caption{Generation of alternative states of an environment: original and sample manipulated semantic maps with relative difference maps.}
\label{fig:sup_maps}
\vspace{-0.2cm}
\end{figure*}



\end{document}